\documentclass[a4paper,10pt]{article}
\usepackage[utf8]{inputenc}
\usepackage{amsmath}
\usepackage[T1]{fontenc}

\usepackage{authblk}

\usepackage[usenames,dvipsnames, table]{xcolor}
\definecolor{green}{RGB}{0,255,0}

\usepackage{hyphenat}
\usepackage{tikz}
\usetikzlibrary{shapes.multipart}

\usetikzlibrary{decorations.markings,plotmarks}
\usetikzlibrary{calc,trees,positioning,arrows,chains,shapes.geometric,%
decorations.pathreplacing,decorations.pathmorphing,shapes,%
matrix,shapes.symbols}
\tikzset{
>=stealth',
punktchain/.style={
rectangle, 
rounded corners, 
draw=black, very thick,
text width=25	em, 
minimum height=3em, 
text centered, 
on chain},
line/.style={draw, thick, <-},
element/.style={
tape,
top color=white,
bottom color=blue!50!black!60!,
minimum width=8em,
draw=blue!40!black!90, very thick,
text width=10em, 
minimum height=3.5em, 
text centered, 
on chain},
every join/.style={->, thick,shorten >=1pt},
decoration={brace},
tuborg/.style={decorate},
tubnode/.style={midway, right=2pt},
}

\tikzset{middlearrow/.style={
decoration={markings,
mark= at position 0.5 with {\arrow{#1}},
},
postaction={decorate}
}
}

\newcommand{\red}[1]{\textcolor{red}{#1}}

\usepackage{bbm}
\usepackage{bbold}
\usepackage[sans]{dsfont}

\usepackage{bm}

\usepackage{color,soul}

\usepackage{url}
\usepackage{hyperref}
\hypersetup{breaklinks=true}
\usepackage{indentfirst}  
\usepackage{array}
\usepackage[normalem]{ulem}
\usepackage[margin=.9in, marginparwidth=.85in,marginparsep=.05in,heightrounded]{geometry}

\usepackage{booktabs}
\usepackage{amssymb, cancel, amsmath, amstext}
\usepackage{epsfig}
\usepackage[small]{caption}
\usepackage{graphicx}
\usepackage[all]{xy}
\usepackage{wrapfig}

\numberwithin{equation}{section}

\usepackage{mathtools}
\usepackage[ruled,vlined]{algorithm2e}

\SetCommentSty{mycommfont}
\usepackage[linguistics]{forest}
\usepackage{qtree}

\usepackage{tikz-cd}
\usepackage[scr=rsfso,cal=zapfc,frak=euler,bb=ams]{mathalfa}

\usepackage{mathrsfs}
\usepackage[shortcuts]{extdash}

\usepackage{stackengine}

\usepackage{caption}
\usepackage{subcaption}
\usepackage{colortbl}
\usepackage{mathrsfs}
\usepackage{multirow}

\usepackage[shortlabels]{enumitem}

\newtheorem{Lemma}{Lemma}[section]

\newtheorem{Remark}[Lemma]{Remark}

{\begin{trivlist} \item[]{\bf Proof. }}%
{\hspace*{\fill}$\rule{.4\baselineskip}{.4\baselineskip}$\end{trivlist}}

\makeatletter
\def\U_#1{U^{\fl{#1}}\@ifnextchar[{\Ubrac}{\relax}}
\def\Ubrac[#1]{#1}
\makeatother

\makeatletter
\def\P_#1{P^{\fl{#1}}\@ifnextchar[{\Pbrac}{\relax}}
\def\Pbrac[#1]{#1}
\makeatother

\makeatletter
\def\W_#1{W^{\fl{#1}}\@ifnextchar[{\Wbrac}{\relax}}
\def\Wbrac[#1]{#1}
\makeatother

\makeatletter
\def\average_#1{\begin{minipage}[t]{1.5cm}
\centering
\emph{Average}\\
\small{$\partial\{#1\}$}
\end{minipage}}
\makeatother

\newcommand{\ma}{_{\text{max}}}
\newcommand{\mi}{_{\text{min}}}

\newcommand{\acst}{\mathscr{A}_i}
\newcommand{\low}{^{\text{(low)}}}
\newcommand{\hi}{^{\text{(high)}}}

\newcommand{\R}{ R}

\newcommand{\ep}{\varepsilon}

\providecommand{\keywords}[1]{\textbf{Keywords:} #1}
\tikzset{
myboxcircle/.style={circle,draw=black,align=center},
}
\tikzset{
myboxrounded/.style={rounded rectangle,draw=black,align=center},
}
\tikzset{
myboxrectangle/.style={rectangle,draw=black,align=center},
}

\definecolor{lightgray}{gray}{0.9}
\definecolor{eggshell}{rgb}{0.94, 0.92, 0.84}
\definecolor{fallow}{rgb}{0.76, 0.6, 0.42}

\definecolor{grannysmithapple}{rgb}{0.66, 0.89, 0.63}

\usepackage{fancyhdr}
\title{Landscape-Sketch-Step: An AI/ML-Based Metaheuristic for Surrogate Optimization Problems}

\author{Rafael Monteiro\thanks{rafael.a.monteiro.math@gmail.com} }
\author{Kartik Sau\thanks{kartik.sau@gmail.com}}
\affil{Mathematics for Advanced Materials Open Innovation Laboratory, AIST, c/o Advanced Institute for Materials Research, Tohoku University, Sendai, Japan.}
\begin{document}

\maketitle

\begin{abstract}

In this paper, we introduce a new  heuristics for global optimization in scenarios  where extensive evaluations of the cost function are expensive, inaccessible, or even prohibitive.
The method, which we call Landscape-Sketch-and-Step (LSS), combines Machine Learning, Stochastic Optimization, and Reinforcement Learning techniques, relying on historical information from previously sampled points to make judicious choices of parameter values where the cost function should be evaluated at. 
Unlike optimization by Replica Exchange Monte Carlo methods, the number of evaluations of the cost function required in this approach  is  comparable to that used by Simulated Annealing, quality that is especially important in contexts like high-throughput computing  or high-performance computing tasks, where evaluations are either computationally expensive or take a long time to be performed. The method also differs from standard Surrogate Optimization techniques, for it does not construct a surrogate model that aims at approximating or reconstructing the objective function.
We illustrate our method by applying it to low dimensional optimization problems (dimensions 1, 2, 4, and 8) that mimic known difficulties of minimization on rugged energy landscapes often seen in Condensed Matter Physics, where cost functions are rugged and plagued with local minima. When compared to classical Simulated Annealing, the LSS shows an effective acceleration of the  optimization process.
\end{abstract}


\section{Introduction}

Minimization problems are ubiquitous in applied sciences, appearing for example in the calibration of machines and power plants \cite{NN_power_plant}, in  combinatorial optimization applications \cite{Cerny},  the optimization  of machine learning models \cite{shalev2014understanding}, and  the prediction of protein folding structures \cite{senior2020improved}. A related - yet different -  problem is that of sampling  low probability events, also referred to as  ``rare''  events \cite[Chapter 8.5]{tuckerman2010statistical}. Such events are remarkably important in Condensed Matter Physics and Biopolymers, fields where rugged energy landscapes - energy potentials, free energies, or simply  cost functions plagued with local minima - are abundant \cite{biopolymers, wales2003energy}. 

The problems we shall focus on  can be formulated mathematically as
\begin{equation}\label{prob_p}
\text{minimize}\quad   E(\theta), \quad \text{for} \quad \theta \in \Omega\subset \R^N, \tag{P}
\end{equation}
under a few hypotheses:
\begin{enumerate}[label= (H\arabic*)]
\item\label{H1}  $\Omega$ is a closed and bounded box.  Such a type of shape domain  is commonly seen in practice by experimentalists performing local searches or adjustments around a given parameter value. 

\item\label{H2}  The cost function (or objective function) $ E(\cdot)$ is continuous on its parameters.  When allied to \ref{H1}, continuity provides sufficient conditions for the existence of a solution to \eqref{prob_p}.  We remark that  further smoothness properties or  bounds on higher order derivatives indicate the cost function's sensitivity on inputs; in this paper such information will not be required, nor used. 

\item\label{H3} The parameter space (or state space) $ \Omega \subset \R^N$ is embedded in a $N$ dimensional space. We shall assume that $N$ is not very high to avoid well-known ``curses'' that happen during sampling in high-dimensional spaces, namely, high-dimensionality exacerbates the  concentration of uniform measures close to  $\Omega$'s boundary \cite[Chapter 2.5]{Hastie}; see also Section \ref{sec:info_sharing}.
\end{enumerate}
Hypotheses \ref{H1}-\ref{H3} are realistic and can be seen in many practical applications, as in the fitting of empirical potentials in Molecular Dynamics \cite{Kartik} or cases where the cost function is not differentiable, rendering inapplicability of  gradient-based optimization methods. 

Several approaches have been used to solve \eqref{prob_p} with varying degrees of success and efficiency. However, most  - if not  all - of them assume that  an indiscriminate number of evaluations  can be made. In contrast, the scenarios we are most interested in - common to many of these practical problems - are those where cost function's evaluation (denoted from now on as $\theta \mapsto  E(\theta)$) is expensive, a situation where indiscriminate exploitation of the objective function $ E(\cdot)$ becomes prohibitive.

The foundation of our study falls on the question of how to use previously sampled elements $\left(\xi,  E(\xi)\right)_{\xi \in \Delta \subset \Omega}$ in order to accelerate the minimization problem. This approach seems reasonable if we assume that the usage of previously sampled data is cheaper than expensive evaluations at any unseen values, or even an opportunity to save computational time and  effort. We have in mind several different cases in which:
\\
\indent (i) Evaluations $\theta \mapsto  E(\theta)$ demands an extensive amount of computations, as required in high-performance computing tasks;\footnote{First principles based molecular dynamics (MD) simulation based on electronic structure are computationally huge expensive, whereas predefined empirical force field based MD simulations are computationally faster \cite{ahuja_rajib}. In this situation, hybrid techniques, where First principles results are used to develop empirical force field, are highly demanding.}
\\
\indent  (ii)  Evaluations $\theta \mapsto  E(\theta)$ take a long time to be completed, as in high-throughput computing tasks; 
\\
\indent (iii) Evaluations $\theta \mapsto  E(\theta)$ can only be obtained after purchase or payment of a consulting fee, at a high cost, as of expert knowledge on a problem. 
\\
It is clear that high evaluation costs play a crucial role in  the architecture of the heuristics proposed in this paper, yet we have no pretension at quantifying - nor constraining - such costs (however, see discussion in Appendix \ref{app:costs}). We will instead guide the discussion to pursue strategies that reduce the objective function evaluations, an idea that we formalize in our last assumption:
\begin{enumerate}[label= (H4)]
\item\label{H4} Let  $\mathscr{H} := \left\{(\xi,  E(\xi)\right\}_{\xi \in \Delta \subset \Omega}$ be a family of parameter-cost function evalution pairs. Then, whenever $\Delta$ is finite, any evaluation $\theta \mapsto  E(\theta)$ on $\theta \not \in \Delta$
is more expensive than interpolations, or inference functions, constructed using the information in $\mathscr{H}$.
\end{enumerate}
Under the hypotheses \ref{H1}-\ref{H4}, we shall refer to  \eqref{prob_p} simply as ``Problem \eqref{prob_p}''. 

In the literature,  optimization methods that advocate for the substitution of a cost function by another -  usually simpler -  one, are commonly known as ``Surrogate Optimization Methods'' and are widely employed in engineering design \cite{forrester2008engineering} and shape optimization \cite{surrogate}. In this paper we shall not use the surrogate modeling approach in the usual way, the main reason being that we do not construct a surrogate function at all. We shall rather tackle the problem by borrowing  several ideas from different fields, like Reinforcement Learning. Therefore, we shall call the pair $\left(\Omega,  E(\cdot)\right)$ an \textit{environment}, a variable that we are optimizing for as an \textit{agent}, and refer to parameter values in $\Omega$ as \textit{states}.

\subsection{A brief overview of some approaches to the minimization problem.}\label{sec:approaches}

In principle, when no further assumptions on Problem \eqref{prob_p} are at hand (for instance, like cost function's convexity), its resolution requires global optimization methods, most of which have a  probabilistic nature. In this section, we explain and contrast a few of them, some of which will serve as stepping-stones to the construction of exploratory policies in the Landscape Sketch and Step approach. 

Let's begin by  discussing Genetic Algorithms (GA), which consist of several agents performing direct searches on parameter space in parallel. The outcome of these searches is gathered and ranked, whereby the  best states are selected and subsequently perturbed. In this way, somewhat mimicking the biological processes of Natural Selection, a new set of states is  generated, and the process starts over.\footnote{In light of that, perturbations are seen as \textit{mutations}.} Although GA is particularly suitable for discrete low-dimensional parameter spaces, it still requires many states to be evaluated at each step, making it very expensive.\footnote{GA should not be dismissed as a valuable technique. In fact, several optimization methods  have been designed based on it; cf. \cite{Hillis}.}  

A different technique is that of  Simulated Annealing (SA), which  has remarkable statistical properties and has been used successfully in Physics, Combinatorial Optimization, among others  applications \cite{Science_sim_ann,SA_bertsekas}. Several objects described next are constructed iteratively, at discrete time steps $i$ called \textit{epochs}. For instance, in this notation, $\theta_i$ indicates the value taken by a parameter $\theta$ at epoch $i$.

In SA's backdrop one needs to introduce a  hyperparameter $T >0$, commonly referred to as \textit{temperature}.  The terminology holds with more intuition in models with Physical meaning and concerns the mobility of atoms, which is proportional to the temperature; this will be further discussed below, but  for our purposes we only emphasize that  $T$ directly impacts how exploratory a policy is.  In practice, one allows a per-epoch dependence $T_i$ usually known as \textit{cooling schedules}, which are mostly  realized  as a   monotonic decay over epochs towards a freezing temperature $T = 0$. For now, we keep the temperature fixed at $T$.

Exploration in Simulated Annealing goes as follows: at first, an initial state $\theta_i^{(T)} \in \Omega$ is randomly perturbed by noise $\xi$ into a neighboring state  $\theta_i^{(T)}  + \xi \in \Omega$; these neighborhoods are designed in order to ensure ergodicity.   Then, the  next state is defined in a  two-steps approach:
\begin{subequations}\label{sim_annealing_transition}
\begin{align}
\widetilde{\theta_i^{(T)}} &= \theta_i^{(T)} + \xi; \label{sim_annealing_transition:a}\\
\theta_{i+1}^{(T)} &= \left\{\begin{array}{ll}
  \widetilde{\theta_i^{(T)}}, & \text{with probability} \quad  \mathrm{exp}\left(\min\left\{\frac{1}{T} ( E(\theta_i^{(T)}) -  E(\widetilde{\theta_{i}^{(T)}}),0\right\}\right);\\
  \theta_i^{(T)},    & \text{otherwise}.   
  \end{array} \right.\label{sim_annealing_transition:b} 
\end{align}
\end{subequations}
In the rest of the paper, we shall call this evolution as \textit{Simulated Annealing with potential} $ E(\cdot)$. This type of approach falls in the class of  Markov Chain Monte-Carlo methods  (MCMC), where  one aims to draw samples from a target distribution by designing a Markov Chain that has the very same target distribution as an invariant measure. In our case, at a fixed temperature $T$, such invariant measure is given by a Gibbs distribution,
\begin{align}\label{Gibbs}
\mu(\theta_i^{(T)}) = \frac{e^{-\frac{1}{T}  E(\theta_i^{(T)} )}}{Z},
\end{align}
where Z is a normalization factor rendering a probability distribution over the parameter states; cf. \cite[Chapter 5.2]{MC}, \cite[Chapter 1]{friedli_velenik_2017}. We remark that,
\begin{enumerate}[label=(\roman*), ref=\theTheorem(\roman*)]
\item  From a Dynamical Systems point of view,  \eqref{sim_annealing_transition} induces a stochastic dynamic on the graph of $ E(\cdot)$ over $\Omega$. (Of course, boundary values must be properly handled; in our case, whenever new points fall out of $\Omega$, we put it  back into the domain by successive reflections accross the boundary.)

\item   At high  temperatures  most transitions are accepted and the landscape $ E(\cdot)$ is mostly ignored. In contrast,  at low temperatures - close to a ``freezing state'' $T \approx 0$ - only  transitions to lower energy states ($ E(\theta_{i+1}) \leq  E(\theta_i)$) are accepted. Thus, one can say that  the policy becomes greedier as the temperature cools down, while more exploratory as the temperature is high.  

\item Going back to the underlying perturbation  \eqref{sim_annealing_transition:a}, if all states are connected (that is, if the underlying Markov chain is "irreducible") then the   dynamics halts at the global minima as $T\downarrow 0$, in what is known as \textit{freezing}; cf. \cite[Chapter 3.1]{Yuval}. However, when  the perturbation is ``too small'', freezing can result in agents getting trapped at local minima due to either insufficient agents' mobility or high-energy barriers; this phenomenon is also known as \textit{tunneling} and  is investigated in Section \ref{sec:examples:toy_model}. In Condensed Matter Physics, such local optima are called \textit{metastable states} (also referred to as ``local optima'').

\item The design of cooling schedules is very important, but almost an art in itself: from the theoretical point of view, they affect mixing properties of Markov Chains and the convergence towards its  invariant measures \cite{hajek1988cooling}; from a practical point of view, if not carefully implemented  they can yield defects in lattice structure and accumulation on metastable states \cite{Science_sim_ann}.
\end{enumerate}
Remark (ii) allows us to borrow some terminology from Reinforcement Learning theory: from now on,  we shall treat cooling schedules as policies and refer to both in an interchangeable way. 

Computationally, Simulated Annealing  relies on extensive  evaluations  $\theta \mapsto E(\theta)$, requires multiple rounds of simulation using ad-hoc cooling schedules, besides trial-and-error initializations. 
One of the biggest challenges of minimization is the acceleration of the search for ground states, which usually requires the development of strategies to overcome metastability issues. Multi-agent exploration is one of the  most effective techniques to overcome this difficulty: it consists of  several agents simultaneously exploring  the environment under different policies (that is, exploration at different temperatures); exploration happens in parallel for a certain number of iterations, after which parallelism is broken. Final states (or ``best states'') are ``combined'' while the information they carry is ``shared'', usually by reassigning states to different agents as initial states for a subsequent round of searches. These ideas have been successfully implemented by Replica Exchange Monte Carlo (REMC)  methods.\footnote{There are several of them, referred to by different names, especially because they have been rediscovered and used in different contexts; cf. \cite{iba2001extended}. Examples are abundant, illustrated by interesting applications in many areas \cite{tokuda_yoshi,biopolymers} } They consist of setting several Simulated Annealing processes  $\Theta_i^{(1)}, \ldots ,\Theta_i^{(m)}$ exploring the environment  under different temperatures; from the point of view of Reinforcement Learning, these processes can be seen as agents exploring the environment under  different policies. In its simplest setting, one defines a set with several states taken by multiple agents at step i,
$$\acst = \left\{\theta_i^{(T_{i}^{(1)})}, \ldots, \theta_i^{(T_i^{(m)})}\right\},$$
where $T_{i}^{(1)} \leq T_{i}^{(2)} \leq\ldots \leq T_i^{(m)}$ and each element corresponds to a different initial condition that will be used in Simulated Annealing exploration, as explained in \eqref{sim_annealing_transition}. 
After $q$ iterations, a permutation $\Pi(\cdot): \{1, \ldots, m \}\to  \{1, \ldots, m \}$ acts on $\mathscr{A}_{i+q}$,  exchanging the states of individuals, 
$$\Pi\left(\mathscr{A}_{i+q}\right) :=\left\{\theta_{i+q}^{(T_{i+q}^{\Pi(1)})}, \ldots, \theta_{i+q}^{(T_{i+q}^{\Pi(m)})}\right\}, $$
which are then  reassigned as initial conditions for future exploration. 
In the end, exploration either halts or a new round of searches is carried out.

The ingenuity REMC method involves the  construction of the permutation $\Pi(\cdot)$ in a probabilistic fashion, obeying a detailed balance condition \cite[Section 3]{iba2001extended}. In this way,  it  allows for  faster mixing and  faster convergence to the ground state. From a Reinforcement Learning perspective, the REMC method is more successful than SA for several reasons, constantly pushing  agents into more exploratory policies at  higher temperatures while, simultaneously, fostering the  adoption of  greedier policies at lower temperatures. In other words, the  exploration-exploitation trade-off is balanced out by adopting exploratory strategies that help in circumventing metastability issues.  The caveat is that the REMC requires many evaluations $\theta \mapsto  E(\theta)$, implying costs proportional to the number of agents and steps required throughout its Simulated Annealing computations.


\subsection{The Landscape Sketch and Step approach}\label{sec:math_settings}
The heuristics suggested in this paper,  which we  call \textit{Landscape Sketch and Step} (in short, LSS), is based on techniques of Reinforcement Learning, Stochastic Optimization, and Machine Learning (ML). Roughly speaking, the LSS approach  is a metaheuristic for global optimization \cite[Section 2]{meta_heuristics} divided into two parts: (i) a ``landscape sketch'' phase, where we construct of a state-value function $\mathcal{V}_i(\cdot)$ (the constructed state-value function at epoch $i$)  and (ii) the design of an exploratory policy that indicates which states shall be taken - or rather, evaluated - next (the ``step'' phase). Exploration is  based on  Simulated Annealing \cite{Aarts}, drawing similarities to  Replica Exchange Monte Carlo methods \cite{iba2001extended} and  Genetic Algorithms \cite{mitchell1998introduction}. 
Both parts happen in tandem and are co-dependent, as is commonly the case in dynamic programming problems.

The function  $\mathcal{V}_i(\cdot)$'s role can be seen from multiple perspectives. From the point of view of Surrogate Optimization, it works as a ``merit function'', an auxiliary tool to indicate candidate points where evaluation should be performed. From the RL point of view, it allows for an informed search strategy, indicating  the most promising  states to be evaluated  by  $\theta \mapsto  E(\theta)$, without any assurance of whether or not  the search's endstate goal has been reached \cite[Section 3.5]{Norvig}. $\mathcal{V}_i(\cdot)$ also  enhances sampling by biasing the dynamics towards ``low probability regions'', yielding accelerated minimization of the cost function; \footnote{These regions are related to the probability  distribution  derived by the Gibbs distribution, constructed from the energy function whose ground state we are searching for.}  this type of biased dynamics pervades the literature on rare events sampling, being present in methods like Hyperdynamics \cite{Hyperdynamics} and Umbrella Sampling, with which the LSS approach shares similarities; cf. \cite{TORRIE,Umbrella_Kast}. In addition to that,  $\mathcal{V}_i(\cdot)$ also plays a similar role to that of a ``cornfield vector'' in Particle Search Optimization: a vector field over parameter space aimed at making particles swarm toward $\theta_i^*$, the best-known parameter at the epoch $i$ \cite[Section 3.2]{PSO_kennedy}. Last, we remark that  when allied  to \ref{H4}, this approach replaces expensive evaluations $\theta \mapsto  E(\theta)$ by cheaper - yet probably inacurate - ones, namely, $\theta \mapsto \mathcal{V}_i(\theta)$.

The construction of $\mathcal{V}_i(\cdot)$ happens over time, using data gathered during exploration of the objective function $ E(\cdot)$. Let's first define $\theta^* := \underset{\theta \in \Omega}{\arg\min}\, E(\theta)$.\footnote{$\theta^*$ could be a set, but this detail is not relevant in the following discussion.}
Ideally, we would like the construction of a state-value function to satisfy two properties,
\begin{align}\label{wishful}
 E(\theta) \geq \mathcal{V}_i(\theta), \quad \forall \, \theta \in \Omega, \quad \mbox{and} \quad \theta^* =\underset{\theta \in \Omega}{\arg\min}\,\mathcal{V}_i(\theta),
\end{align}
where $\mathcal{V}_i(\cdot)$ is a ``nice'' function, possibly convex.  However, $\theta^*$ is the very (unknown) object we are searching for. Hence, we are forced to ``relax'' \eqref{wishful} by restricting the parameter $\theta$ to the set  $\mathscr{H}_i$ containing states previously sampled up to the epoch $i$, 
\begin{align}\label{history}
\mathscr{H}_i := \left\{\xi \in \Omega \,\Big\vert \,  E(\xi)\, \text{has been evaluated and stored up to time step $i$}\right\}.
\end{align}
Thus, if we define\footnote{In the Surrogate Optimization literature, $\theta_i^*$ is called ``the incumbent'' point.}
\begin{align}\label{history_theta}
\theta_i^* := \underset{\xi \in \mathscr{H}_i}{\arg\min}\, E(\xi)
\end{align}
we now  aim instead at satisfying
\begin{align}\label{wishful_relaxed}
 E(\theta) \geq \mathcal{V}_i(\theta), \quad \forall \, \theta \in \mathscr{H}_i, \quad \mbox{and} \quad \underset{\xi \in \mathscr{H}_i}{\arg\min}\,\mathcal{V}_i(\xi) = \theta_i^*.
\end{align}
These properties allow us to introduce the quantity
\begin{align}\label{M_functional}
i \mapsto\mathcal{M}_i :=  E(\theta_i^*), 
\end{align}
which refer to as the \textit{M-functional}, a quantity that  is monotonically decaying over time thanks to the second property in \eqref{wishful_relaxed} and to the fact that $\mathscr{H}_i$ are nested sets over epochs. From a purely quantitative point of view,  we use the $M$-functional \eqref{M_functional} instead of  $ E(\cdot)$  to avoid possible oscillatory behavior of the latter; indeed, for any $\Delta \subset \mathscr{H}_i$, the quantity
$\displaystyle{\max_{\theta \in \Delta} E(\theta) - \min_{\theta \in \Delta} E(\theta)}$
tends to be highly oscillatory throughout exploration, since the LSS approach fosters exploration by always stepping into non-optimal states. Evidently, by definition, it holds that
\begin{align*}
 E(\theta) \geq \mathcal{M}_i, \quad \text{for all} \quad \theta \in \mathscr{H}_i;
\end{align*}
our goal is to ensure that the limit $\displaystyle{\lim_{i\to \infty} E(\theta_i^*) = \lim_{i\to \infty}\mathcal{M}_i =  E(\theta^*)}$ is attained. 

\begin{figure}[htbp]
\centering
\includegraphics[width=.8\textwidth]{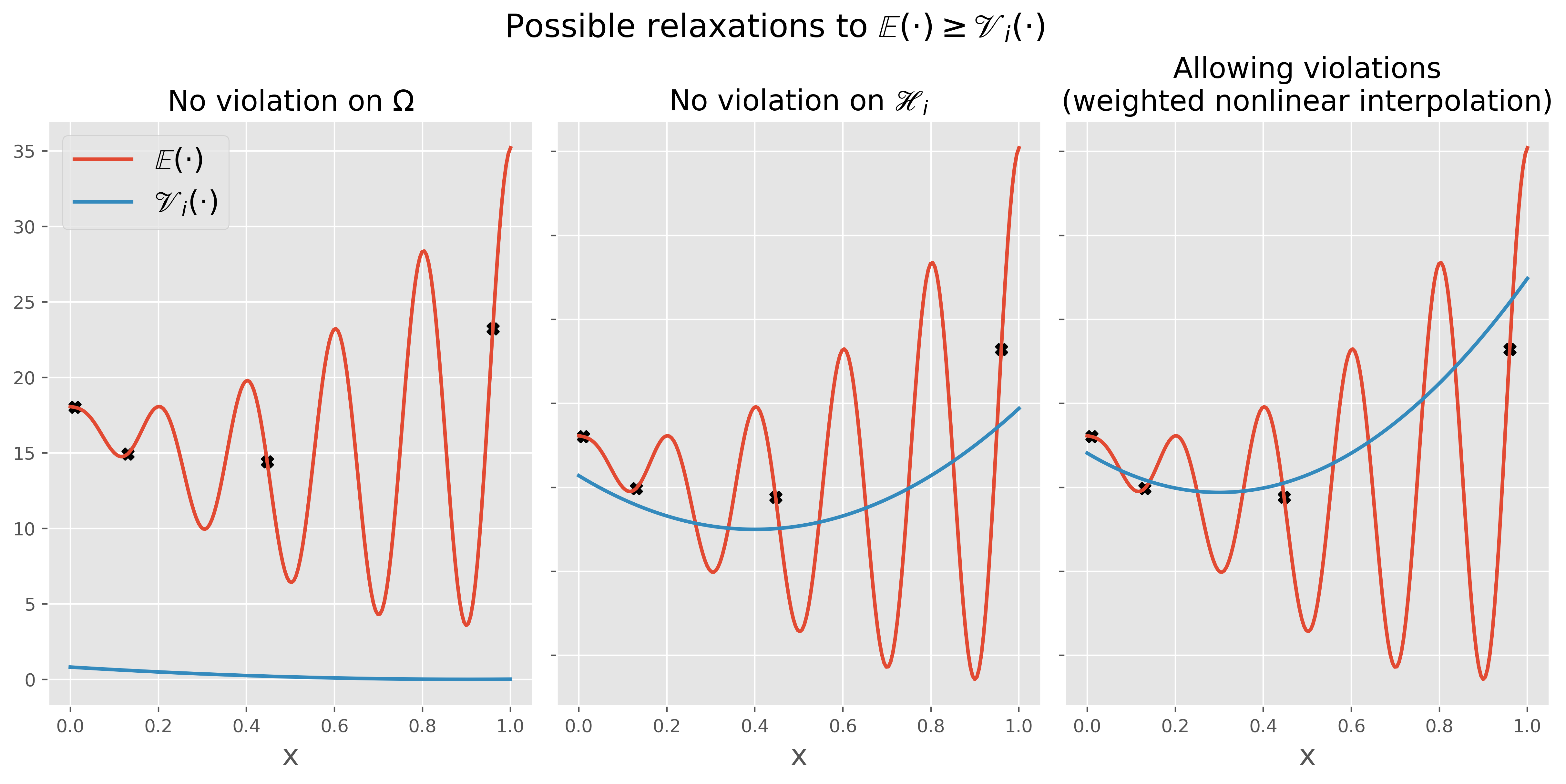}
\caption{ The construction of the state-value function in different contexts, according to the inequality $ E(\cdot) \geq \mathcal{V}_i(\cdot)$. On the left, the conditions in \eqref{wishful} are fully satisfied but, in practice, unfeasible due to partial information about the environment. In the middle, the conditions in \eqref{wishful} are relaxed into new condition \eqref{wishful_relaxed}, which are fully verified on sampled points (black markers). On the right figure, weights are used to assign more relevance to states in $\mathscr{H}_i$ that take evaluated values $\theta \mapsto  E(\theta)$ closer to $\mathcal{M}_i$; both conditions in \eqref{wishful_relaxed} are violated. \label{fig:violations}}
\end{figure}

Although \eqref{wishful_relaxed} can be trivially satisfied (for instance, $\mathcal{V}_i(\cdot) :=  E(\theta_i^*)$ would do),  or even assured with careful reweighing of parameters, it is convenient to adopt an even more relaxed condition. In fact,  we shall use the set $\left\{(\xi,  E(\xi)|\, \xi \in \mathscr{H}_i\right\}$  for interpolation with  weights $\mathcal{W}_i(\cdot)$, biasing values  of $ E(\cdot)\big|_{\mathscr{H}_i}$  close to $\mathcal{M}_i$; in this manner, we shall expect but accept some violations of both properties in \eqref{wishful_relaxed}. The trade-off is clear: by deliberately allowing violations in \eqref{wishful_relaxed}, we may compromise the dynamics but allow for automated reweighting,  an evident advantage, especially when contrasted with ad-hoc weight construction (an issue that other  methods, like hyperdynamics, suffer from   \cite[Section 4]{TORRIE}).\footnote{It must be said that the term bias has a dynamical system connotation rather than a statistical meaning. Indeed, we are not interested at constructing $\mathcal{V}_i(\cdot)$ as a biased estimator to $ E(\cdot)$.} 
All these properties are illustrated in Figure \ref{fig:violations}. 

There are several differences between the LSS and other Surrogate Optimization methods. The first and most important of them is the absence of a surrogate model: $\mathcal{V}(\cdot)$ plays instead the role of a merit function, being used with the sole purpose of choosing from candidate points where further evaluations $\theta \mapsto  E(\theta)$ will be carried out. Second, the generation of candidate points by branching from previous points, a process driven by Simulated Annealing nature with respect to the $\mathcal{V}(\cdot)$. The third idea is that of weighting elements not only by their relevance (when compared with $\mathcal{M}_i$), but also with how recently it has been introduced to the queue of active states. This also allows exploration to follows a multi-agent nature, obeying different exploratory policies, which seems more suitable to address the multi-scale nature of many of the corrugated cost functions seem in practice. Although designs of weights that take into account distance to previously evaluated points are seem in the literature, as in  \cite{wang2014general}, they do not account for th

At each epoch $i$, exploration is carried out by using a subsample  $\acst$  elements of elements in $\mathscr{H}_i$ whose role  will be further  explained later. In a nutshell,  the LSS approach at epoch $i$ goes as follows:

\begin{algorithm}[H]\label{alg:LSS}
\SetAlgoLined
\SetInd{.2cm}{.2cm}
\KwResult{Find a candidate for minimizer of Problem \eqref{prob_p}. }
\KwData{Start with a set $\mathscr{H}_0 \subset \Omega$ as in \eqref{history} and a halting condition.}
\While{The halting condition has not been reached}
{
\Indp (i) Use $\mathscr{H}_i$ to construct $\mathcal{V}_i(\cdot)$ by weighted nonlinear interpolation;
\vspace{.1cm}

(ii) Judiciously select a subsample $\acst$ of  $\mathscr{H}_i$ with a pre-defined size. For  each element in $\acst$ (an agent), perform Simulated Annealing with potential $\mathcal{V}_i(\cdot)$, under $k$ different temperatures $T^{(1)} < \ldots <T^{(k)}$ (see Section \ref{sec:approaches}). The final states in each of these simulations make up the  sets  $\acst^{(T^{(1)})}, \ldots , \acst^{(T^{(k)})}$;
\vspace{.1cm}

(iii) Following a predefined rule that defines the maximum number of costly evaluations per epoch, select a few states among $\acst^{(T^{(1)})}, \ldots , \acst^{(T^{(k)})}$. Each element not yet in $\mathscr{H}_i$ will be evaluated through  $\theta \mapsto  E(\theta)$;
\vspace{.1cm}
(iv) Aggregate new evaluations to the history set $\mathscr{H}_i$, now named as $\mathscr{H}_{i+1}$;
\vspace{.1cm}
(v) With the sampled elements in hands and following a pre-defined policy, evolve the set $\acst$ to $\mathscr{A}_{i+1}$;
}

\caption{The LSS approach, in algorithmic format.}
\end{algorithm}

Applications of the Algorithm \ref{alg:LSS} are broad and go beyond the heuristics suggested in this paper. For instance, if at step (v) one simply defines $\mathscr{A}_{i+1}$ as a permutation of $\acst$ realized by the action of a detailed balance law that enforces ergodicity, one obtains a variation of the REMC method; see discussion in Section \ref{sec:discussion}.

Several remarks are readily available. First, and above all, let's highlight that Simulated Annealing in step (ii) is carried out with potential $\mathcal{V}_i(\cdot)$, not $ E(\cdot)$. In a certain way, we can think of $\mathcal{V}_i(\cdot)$ as an emulated environment, whereas $ E(\cdot)$ is the real one. This replacement is key to reducing expensive evaluations $\theta \mapsto  E(\theta)$. Furthermore, thanks to \ref{H4}, we precede expensive evaluations $\theta \mapsto  E(\theta)$ by several cheaper evaluations $\theta \mapsto \mathcal{V}_i(\theta)$ with the intention of  accelerating minimization (that is, faster decay of $ E(\theta_i^*)$). 

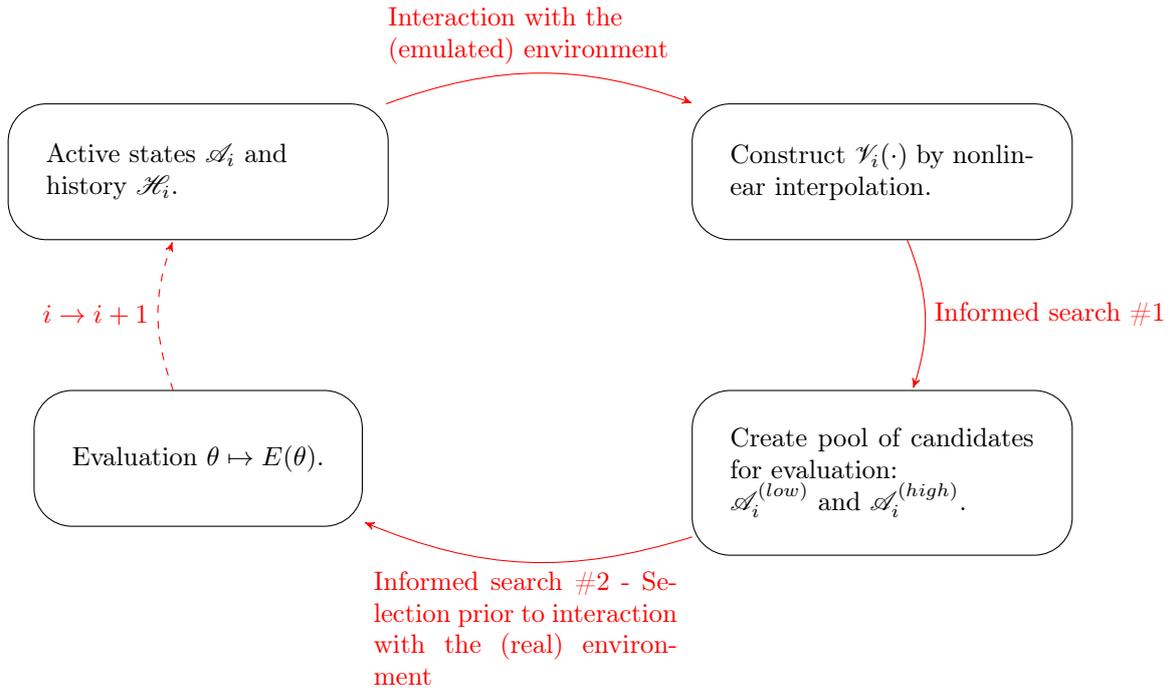
\begin{figure}[htbp]
\centering 
\tikzset{
root/.style={circle,draw=red!70,fill=red!30},
leaf/.style={circle,draw=blue!70,fill=blue!30},
label/.style={sloped,above}
}
\begin{tikzpicture}[node distance=2cm,
mynode/.style={
    inner sep=.5cm,
    minimum size=1.8cm,
    outer sep=0pt,
    clip,
    rounded corners=0.5cm,
    draw
},
myarrow/.style={
    ->,
    red,
    shorten >=1pt,
    bend angle=20,
    bend left
},
dashedarrow/.style={
    ->,
    dashed,
    red,
    shorten >=1pt,
    bend angle=20,
    bend left
}
]
%
\node[mynode] (A) {\begin{minipage}{4cm}
		Active states $\acst$ and \\history $\mathscr{H}_i$. 
		\end{minipage}
		};
\node[mynode, right=4cm of A] (B) {\begin{minipage}{4cm}
				Construct $\mathscr{V}_i(\cdot)$ by nonlinear interpolation.
				\end{minipage}};
\node[mynode, below=of B] (C) {\begin{minipage}{4cm}
			    Create pool of candidates for evaluation:\\ $\acst^{(low)}$ and $\acst^{(high)}$.
			    \end{minipage}};
\node[mynode, below=of A] (D) {Evaluation $\theta \mapsto  E(\theta)$.};

\draw[myarrow] (A) to node [above] {
\begin{minipage}{4cm}
\red{Interaction with the } \\\red{(emulated) environment} 
\end{minipage}
} (B);
\draw[myarrow] (B) to node [right] {\red{Informed search \#1}} (C);
\draw[myarrow] (C) to node [below] {
\begin{minipage}{4cm}
\red{Informed search \#2 -  Selection prior to interaction with the (real) environment} 
\end{minipage}} (D);
\draw[dashedarrow] (D) to node [left] {\red{$i \to i+1$}} (A);
\end{tikzpicture}
\caption{An overview or the exploration-exploitation approach in the Landscape-Sketch-Step heuristics. Observe that if  we skip the two nodes on the right we are simply describing the Simulated Annealing method. \label{fig:diagram}}
\end{figure}

As said above, the framework of the LSS approach is broad. Yet, to simplify our exposition, throughout the paper we use only two cooling schedules, represented as $T_i^{\text{(low)}} <T_i^{\text{(high)}}.$ Thus, we  represent the extended set $\acst^{(T_i^{\text{(low)}})}$ as $\acst^{\text{(low)}}$ and $\acst^{(T_i\hi)}$ as  $\acst\hi$. In this scenario,  the LSS approach can be depicted by the diagram in Figure \ref{fig:diagram}.

As depicted in Figure \ref{fig:diagram}, the construction of $\mathcal{V}_i(\cdot)$ is followed by that of $\acst\low$ and $\acst\hi$. In between, an informed search takes place, where Simulated Annealing  with potential $\mathcal{V}_i(\cdot)$ is carried out. During one epoch cycle the number of steps used to construct $\acst\low$ and $\acst\hi$ is fixed, but can vary over different epochs or even follow its own schedule; for instance, the number of steps can be small in the initial epochs and grow towards the end. Furthermore, over epochs the parameter space $\Omega$ can be shrunk into smaller domains $\Omega_i \subset \Omega$, an evolution that can be controlled by hyperparameters and is referred to as \textit{box-shrinking schedules}. Therefore, since parameter search happens inside $\Omega_i$, we shall call this informed search part of an epoch cycle a  \textit{within-box search steps}; for more details, see Section \ref{sec:temperature}. 

Finally,  let's clarify the role of $\acst$, which is a subsample of $\mathscr{H}_i$ known to us as the \textit{queue of active states}, or simply, as \textit{active states}. These are states from which further exploration of the environment $\left(\Omega_i,  E(\cdot)\right)$ will be carried out. As a queue, $\acst$ is endowed with a data structure property: it is a sequence that can grow in size while  obeying a "First-In-First-Out" property, that is, the first element added to it is also the first that can be removed. Its elements are tagged by ordered indexes and two operations are allowed: an \textit{enqueueing} operation, by which elements are inserted into $\acst$ with an indexed tag strictly greater than those already in the queue (if the queue is empty, the element takes the index zero), and (whenever non-empty) a \textit{dequeueing} operation, by which the lowest indexed element is removed from it; cf. \cite[Chapter 10]{Cormen}. In the context of the LSS, we observe that even though  $\acst$ is not endowed with any thermodynamic interpretation, nor it is  a ``low dimensional'' representation of $\mathscr{H}_i$, it  plays an analogous role to collective variables in Metadynamics; cf. \cite{Laio_2008, tuckerman2010statistical}. 

Through the lenses of Reinforcement Learning, one can say that the LSS approach has a  multi-agent nature, where  states assumed by agents are our main interest; in fact, active states $\acst$ aggregates the parameters we are optimizing for: under a given cooling schedule policy $T_{i}$, each element in the queue $\acst$ is the starting point of an exploratory  realization of a Simulated Annealing process with potential $\mathcal{V}_i(\cdot)$  which, after  a predefined number of steps, yields  a new sample $\acst^{(T_{i})}$ of elements in $\Omega_i$. When several different exploratory policies $T_{i}^{(1)}, \ldots, T_i^{(m)}$ are given, an equivalent number of final samples $\acst^{(T_{i}^{(1)})}, \ldots, \acst^{(T_{i}^{(m)})}$ are obtained. The process of selecting which among the elements in $\acst^{(T_{i}^{(1)})}\cup \ldots \cup \acst^{(T_{i}^{(m)})}$ are ``worth'' to be evaluated $\theta \mapsto  E(\theta)$ is defined in a probabilistic fashion, with a strong bias toward values that are lower or close to $\mathcal{M}_i =  E(\theta_i^*)$; see Section \ref{sec:info_sharing}.

It should be mentioned  that even though we are discussing an optimization problem, along the way, it will be  more convenient to adopt terminology from Dynamical Systems and Reinforcement Learning (RL). There are several reasons for doing this.

First,  because  the  search for a ground state (global minimum) takes place dynamically. Such search requires hyperparameter adjustment along the way,  changes that must happen in an  automatic fashion  due to agents interacting with the environment, learning, finding out strategies to better explore it and, consequently, tuning some hyperparameters accordingly. 

Second, the multi-agent nature of the LSS approach implies that $\acst$ is comprised of several active states that are used concomitantly.  Throughout exploration, active states gather and share information  about the environment, while consensus guides the selection of the best among them.

Last, consensus also drives the decision of where and how new evaluations $\theta \mapsto  E(\theta)$ are made; that is the main reason behind the  construction of a state-value function. This will  help us to carry out informed searches by assigning values to regions in the parameter space, allowing extrapolation  - i.e.,  generalization -  to states yet unseen. In this regard,  we will borrow some ideas from Artificial Intelligence and  Reinforcement Learning.

\subsection{Related work}\label{sec:related}

In recent years, there has been a steady growth in the computational capabilities and efficiency of Deep Learning toolboxes; nowadays, it is not uncommon to see industrial state-of-the-art models with millions - if not billions - of hyperparameters. Naturally,  optimization of these models with respect to their hyperparameters has become the focus of extensive research \cite{bergstra2011algorithms, White_Neiswanger_Savani_2021}. Some of the outcomes of this fast development are spread across many scientific fields, especially those requiring rare events sampling, like protein folding in Biology \cite{jumper2021highly} and Molecular Dynamics in Physics  \cite{schutte_klus_hartmann_2023,NOE202077}. The intricate multiscale nature of these phenomena is quite salient and which approach is best or the most efficient to handle such difficulties is still the focus of research \cite{bonati2021deep}; the LSS approach, using multiple agents, is one among them.

The topic of multi-agent systems is discussed extensively in \cite{panait2005cooperative}, where the authors avoid a single definition of the term; due to that,  we mean ``multi-agent exploration'' in a broad sense; one in which independent agents cooperate with respect to a joint reward, adapting their behavior based on information gathered by the group. We emphasize that in the LSS there is not a master-slave situation with respect to agents (that is, the behavior of one defines the behavior of others). However, it is clear that $\theta_i^*$ is treated asymmetrically, for new agents are more prone to branch out from it. We point out the exciting discussion about communication between agents carried out in  \cite[Section 4]{panait2005cooperative}, arguing that some methods can be orthogonalized, and several agents substituted by a single one. We emphasize that this is not the case of the LSS, for the independence of exploration by Simulated Annealing is intercalated with information gathering and the construction of $\mathcal{V}(\cdot)$.

As explained in Section \ref{sec:math_settings}, ML techniques are very suitable in Surrogate Optimization problems. 
Indeed,  the sequential data gathering in contexts of high-cost objective functions has pushed the development of techniques where surrogate models quickly adjust to the available information, as for instance in Bayesian Optimization \cite{frazier2018tutorial, NIPS2012_05311655} or, as in other contexts, through RL models that learn the environment through online data \cite[Chapter 8]{RL}.  Our use of a queue of active states is inspired by these ideas.

With concern to the applications of these techniques in combination, the paper \cite{ilievski2017efficient} discusses interesting problems in hyperparameter optimization where, like in Problem \ref{prob_p}, a black box model has its parameters optimized on a constrained domain. After a thorough discussion of inefficiencies in classical surrogate optimization models, they propose a new hyperparameter optimization method that uses RBF to approximate error surrogates, refining and improving ideas in other papers by some of the authors \cite{regis2007stochastic,wang2014general}. As in \cite{wang2014general},  the authors address the problem of exploration and exploitation through a weighted score $W(\cdot)$  that balances surrogate estimates in the vertical direction $V^{ev}(\cdot)$ and in terms of the distance to previously sampled points $V^{dm}(\cdot)$ in a convex combination fashion $W(\cdot) = wV^{ev}(\cdot) + (1-w)V^{dm}(\cdot)$, cycling the variable $w$ on a discrete subset of $[0,1]$. Unlike in our study, they do not link the role of $w$ to the idea of agents` spatial concentration, nor to any quantifier that captures global information about the system; cf. Section \ref{selection_measures}.


%

\paragraph{Outline of the paper.} In Section  \ref{sec:math_comp_stat} we describe the theoretical background of the model; in Section,\ref{sec:examples} we describe a few applications, mostly inspired by Material Sciences problems; finally, in Section \ref{sec:discussion}, we summarize our results and indicate future directions of investigation.

\paragraph{Notation.} The set $ R^N$ denotes the real-valued, $N$ dimensional Euclidean space.  We say that $f(\cdot):  R \to  R$ is $\mathscr{O}(N)$ if there exists a $C>0$ such that $\vert f(N) \vert \leq C \vert N\vert.$ When $\mathscr{B}$ is a set (respectively, a queue), $\vert \mathscr{B}\vert$ denotes its cardinality (respectively, its length). Last, for any two queues  $\mathscr{A} = (a_1, \ldots, a_n)$ and $\mathscr{B} = (b_1, \ldots, b_m)$, we define $\mathscr{A}\oplus \mathscr{B} := (a_1, \ldots,a_n, b_1, \ldots, b_m)$.

\section{Mathematical, computational, and statistical background}\label{sec:math_comp_stat}

In this section  we describe in detail the construction and usage of $\mathcal{V}_i(\cdot)$, a function used as a state-value function. It is used  not as a mean to approximate $ E(\cdot)$, but rather to indicate regions in $\Omega_i$ that are more promising for further searching, prior to evaluations $\theta \mapsto  E(\theta)$. This is in sharp contrast with the methods described in Section \ref{sec:approaches} (none of which have an inference step), being cheap thanks to \ref{H4}.

\subsection{Information  gathering and construction of a state-value type function -- biasing the dynamics towards low probability regions}\label{sec:info_gathering}

As described in Section \ref{sec:math_settings}, the construction of  $\mathcal{V}_i(\cdot)$ is deeply tied to the set $\mathscr{H}_i$ and can be carried out using any nonlinear interpolation method that  weighs data points, aiming at the satisfaction of the inequalities in \eqref{wishful_relaxed} but possibly violating them. There are two auxiliary functions in the backdrop of this construction,   weights $\mathcal{W}_i(\cdot)$ and a ranking-type function $\mathcal{R}_i(\cdot)$,  both of which evolve through epochs, somehow  in tandem with the insertion of new elements into $\mathscr{H}_i$. From the point of view of RL, $\mathcal{R}_i(\cdot)$ plays the role of a reward function.

Recalling the definition of $\theta_i^*$ in \eqref{history_theta}, we begin by explaining $\mathcal{R}_i(\cdot)$, a function used to rank elements in  $\theta \in \mathscr{H}_i$ by assignning higher ratings to evaluations $ E(\theta)$ closer to $\mathcal{M}_i =   E(\theta_i^*)$. It is defined as
\begin{align}
\mathcal{R}_i(\theta) = e^{-\beta \left( E(\theta) - \mathcal{M}_i\right)},
\end{align}
where $\beta$ is a normalization factor that depends on hyperparameters. Clearly,  $\mathcal{R}_i(\cdot)$ takes values in the range $[0,1]$ in such a way that $\mathcal{R}_i(\theta_i^*) = 1$.

Next, we define a weight function $\mathcal{W}_i(\cdot)$ over $\mathscr{H}_i$ by  taking into account whether  $\xi$ has been evaluated before or not: if it has, we assing an initial weight $\mathcal{W}_i(\xi) = 1$, otherwise a value $\mathcal{W}_i(\xi)$ already exists. 
Once this step has finished, $\mathcal{W}_i(\cdot)$ evolves according to a temporal-differencing scheme,
\begin{align}\label{weigtt_temp_diff}
\mathcal{W}_i(\theta)\gets \mathcal{W}_i(\theta) + \alpha\left(\mathcal{R}_i(\theta) - \mathcal{W}_i(\theta)\right), \quad \text{for} \quad \theta \in \mathscr{H}_i,
\end{align}
where $\alpha \in [0,1]$ is a hyperparameter. Observe  that $\mathcal{W}_i(\theta) \in [0,1]$ for all $\theta \in \mathscr{H}_i$. Furthermore,   $\mathcal{W}_i(\theta_i^*) =  1$ is an  invariant of the above construction. 

Note that initially \eqref{weigtt_temp_diff} assigns more weight  to states that have been recently added to $\mathscr{H}_i$, inducing more exploration in the beginning, but adjusting this weight over time by the ranking function $\mathcal{R}_i(\cdot)$. Moreover, $\mathcal{R}_i$ weights favorably lower values of $ E(\cdot)$. 

Once weights $\mathcal{W}_i(\cdot)$ are available, we construct the state-value-type function $\mathcal{V}_i(\cdot)$ by nonlinear interpolation using a ML model of Supervised Learning type on the set of  triplets $\left\{\left( \widetilde{\theta},  E(\widetilde{\theta}), \mathcal{W}_i(\widetilde{\theta})\right)_{\widetilde{\theta} \in \mathscr{H}_i}\right\}$. 

Finaly, the construction of $\mathcal{V}_i(\cdot)$ by interpolation takes place at the beginning of epoch cycles, every k-epochs, where $k$ is a hyperparameter.  

\subsubsection{Computational effort to contruct $\mathscr{H}_i$ - choosing among different ML model architectures.}\label{comput_effort}

Different ML models imply different qualities  of the functional space used for interpolation. Prior to any computation, two issues need to be contemplated:

\begin{enumerate}[label=(\roman*), ref=\theTheorem(\roman*)]

\item The ML model should scale well on the size of the history. For instance,  Support Vector Regression (SVR) with nonlinear kernels scales as $\mathcal{O}(K^2)$ or $\mathcal{O}(K^3)$ on an input with size $K$, which is prohibitive. For that reason we applied the LSS approach with ML models that scale well on the input size: SVR with linear kernels and Artificial Neural Networks.

\item Despite the model chosen, the LSS approach is constantly collecting a growing history of measurements due to agents' interaction with the environment, a history that is fed to the ML model. Hence,  the ML  model evolves with a tendency to overfit, especially if a thorough sweep and evaluation over the domain  $\Omega_i$ is carried out. For that reason, it is important to find ways to avoid overfitting by pushing the model away from obsessively searching regions that have already been well explored. We achieved that by selecting ML models among many different architectures or hyperparameters using a multi-armed bandits method with an $\ep$-policy \cite[Chapter 2]{RL}, forcing several models to be trained from time to time and keeping a score for each of them, which	 is updated by temporal differencing as to lower the importance of earlier fittings.  Whenever a multi-armed bandits technique is not used, we select, evaluate, and compare different ML models at every $k$-epochs, selecting the best among them; the LSS approach sticks to the best one among them for the next $k$ epochs.\footnote{In the code, we associate these $k$-epochs intervals with different boxes $\Omega_i \subset \Omega_i$, that can also shrink in size. We call these changes \textit{box shrinking schedules}.  This adds flexibility to the parameter search, for boxes may also be designed to shrink over time, reducing the size of the parameter space. For further details, see \cite{MD_ML_github}.}
\end{enumerate}
As more information is gathered through exploration, the history $\mathscr{H}_i$ evolves and gets larger. Still, the question of how to find these points  and enhance exploration is yet to be clarified and  will be explained next.

\subsection{Information sharing and construction of  exploratory policies -- reaching out the consensus of which states should be evaluated next}\label{sec:info_sharing}

Extending $\mathscr{H}_i$ involves choosing new elements in $\Omega_i$ that will be used to improve exploration of the environment. This is done by selecting a  subset  $\acst$ - referred to as the \textit{queue of active states} at epoch $i$ - whose construction obeys the following invariants,
\begin{align}\label{property_min}
\acst \subset \mathscr{H}_i,\footnotemark[9]{} \quad \text{and} \quad \min_{\theta \in \acst} E\left(\theta\right) =\min_{\theta \in \mathscr{H}_i} E\left(\theta\right).
\end{align}
\footnotetext[9]{ By this we mean that the elements in the queue $\acst$ are in the set $\mathscr{H}_i$; we shall adopt this abuse of notation unapologetically throughout the rest of the paper. }\addtocounter{footnote}{+1}
As explained in Algorithm \ref{alg:dynamics_Hi} the dynamics of $\acst$ precedes and helps to explain that  of $\mathscr{H}_i$,  guaranteeing that the first property in \eqref{property_min} holds.

\begin{algorithm}[H]\label{alg:dynamics_Hi}
\SetAlgoLined
\KwResult{At epoch $i+1$, construct the history set  $\mathscr{H}_{i+1}$ }
\KwData{$\theta \in  \mathscr{A}_{i+1}$, $\acst\subset \mathscr{H}_{i}$, and $\mathscr{H}_{i}$.}

\eIf{$\theta \in \acst$}{
\tcc{In this case, $\theta \in \acst\subset \mathscr{H}_{i}$}

$\mathscr{H}_{i+1} : = \mathscr{H}_{i}$;
}{
\tcc{In this case, $\theta \in \mathscr{A}_{i+1}\setminus \acst$}

Evaluate  $\theta \mapsto  E(\theta)$ and store $ E(\theta)$;

$\mathscr{H}_{i+1}\gets \mathscr{H}_{i} \cup \{\left(\theta,  E(\theta)\right)\}$.
}

\caption{The dynamics of $\mathscr{H}_i$}
\end{algorithm}

Thus, it remains to explain how $\acst$ evolves, a process in great part controlled by a few hyperparameters, $K_i\low$, $K_i\hi$,  $e_i$, and $a_{i+1}$, each one of them explained in detail  below.\footnote{In all the simulations performed, the sequences $e_i$ and $a_i$ satisfies $e_i \leq a_i$ (otherwise elements would be evaluated but not stored into $\mathscr{H}_i$) and were non-increasing. We assume that $\vert\mathscr{H}_0\vert\geq a_0$. If not, we sample $\Omega_i$ several times, until the selection process is possible.} The process essentially unfolding  in three stages: selection, enlargement, and trimming.

\begin{figure}[htbp]
\centering
\begin{subfigure}[b]{0.46\textwidth}
\begin{tikzpicture}[>=latex]
\draw (0,0) node {
	\begin{minipage}{3cm}
	\centering
	  $\left( \theta_1, \theta_2, \theta_3 \right)$\\
	  \vspace{.1cm}
	  $\acst$
	\end{minipage}};
\begin{scope}[shift={(-2,-2)},rotate=30]
\draw (0,0) node {	    
\begin{minipage}{3cm}
	\centering
	  $\acst\low$\\
	  \vspace{.1cm}
	  $\left( \red{\theta_1\low}, \theta_2\low, \theta_3\low \right)$
	\end{minipage}};
\end{scope}

\begin{scope}[shift={(2,-2)},rotate=-30]
\draw (0,0) node {
\begin{minipage}{3cm}
	\centering
	  $\acst\hi$\\
	  \vspace{.1cm}
	  $\left( \theta_1\hi, \red{\theta_2\hi}, \theta_3\hi \right)$
	\end{minipage}};
\end{scope}  
\draw[->, thick] (0,-0.5) -- (2,-1.3);
\draw[->, thick] (0,-0.5) -- (-2,-1.3);

\draw (0,-3) node {
$\theta_i^* = \theta_1$};
\end{tikzpicture}
\caption{}
\end{subfigure}
\qquad
\begin{subfigure}[b]{0.46\textwidth}
\begin{tikzpicture}[>=latex]
\draw (0,0) node {
	\begin{minipage}{3cm}
	\centering
	  $\left( \theta_1, \theta_2, \theta_3 , \red{\theta_1\low}, \red{\theta_2\hi}\right)$\\
	  $\widetilde{\acst}$
	  \vspace{.2cm}
	\end{minipage}};
\begin{scope}[shift={(-2,-2)},rotate=30]
\draw (0,0) node {	    
\begin{minipage}{3cm}
	\centering
	  $\acst\low$\\
	  \vspace{.1cm}
	  $\left( \red{\theta_1\low}, \theta_2\low, \theta_3\low \right)$
	\end{minipage}};
\end{scope}
\begin{scope}[shift={(2,-2)},rotate=-30]
\draw (0,0) node {
\begin{minipage}{3cm}
	\centering
	  $\acst\hi$\\
	  \vspace{.1cm}
	  $\left( \theta_1\hi, \red{\theta_2\hi}, \theta_3\hi \right)$
	\end{minipage}};
\end{scope}
\draw[->, thick] (0,-0.5) -- (2,-1.3);
\draw[->, thick] (0,-0.5) -- (-2,-1.3);

\draw (0,-3) node {
$\theta_i^* = \theta_1$};
\end{tikzpicture}
\caption{}
\end{subfigure}
\vspace{1cm}
\begin{subfigure}[b]{0.46\textwidth}
\begin{tikzpicture}[>=latex]
\draw (0,0) node {
	\begin{minipage}{3cm}
	\centering
	  $\left(\theta_3 , \red{\theta_1\low}, \red{\theta_2\hi}\right)$
	  \vspace{.1cm}
	  $\widetilde{\acst}$
	\end{minipage}};
\begin{scope}[shift={(-2,-2)},rotate=30]
\draw (0,0) node {	    
\begin{minipage}{3cm}
	\centering
	  $\acst\low$\\
	  \vspace{.1cm}
	  $\left( \red{\theta_1\low}, \theta_2\low, \theta_3\low \right)$
	\end{minipage}};
\end{scope}
\begin{scope}[shift={(2,-2)},rotate=-30]
\draw (0,0) node {
\begin{minipage}{3cm}
	\centering
	  $\acst\hi$\\
	  \vspace{.1cm}
	  $\left( \theta_1\hi, \red{\theta_2\hi}, \theta_3\hi \right)$
	\end{minipage}};
\end{scope}
\draw[->, thick] (0,-0.5) -- (2,-1.3);
\draw[->, thick] (0,-0.5) -- (-2,-1.3);

\draw (0,-3) node {
$\theta_i^* = \theta_1$};
\end{tikzpicture} 
\caption{}
\end{subfigure}
\qquad
\begin{subfigure}[b]{0.46\textwidth}
\begin{tikzpicture}[>=latex]
\draw (0,0) node {
	\begin{minipage}{3cm}
	\centering
	  $\left(\red{\theta_1\low}, \red{\theta_2\hi}, \theta_1\right)$
	  \vspace{.1cm}
	  $\mathscr{A}_{i+1}$
	\end{minipage}};
\begin{scope}[shift={(-2,-2)},rotate=30]
\draw (0,0) node {	    
\begin{minipage}{3cm}
	\centering
	  $\acst\low$\\
	  \vspace{.1cm}
	  $\left( \red{\theta_1\low}, \theta_2\low, \theta_3\low \right)$
	\end{minipage}};
\end{scope}
\begin{scope}[shift={(2,-2)},rotate=-30]
\draw (0,0) node {
\begin{minipage}{3cm}
	\centering
	  $\acst\hi$\\
	  \vspace{.1cm}
	  $\left( \theta_1\hi, \red{\theta_2\hi}, \theta_3\hi \right)$
	\end{minipage}};
\end{scope}
\draw[->, thick] (0,-0.5) -- (2,-1.3);
\draw[->, thick] (0,-0.5) -- (-2,-1.3);

\draw (0,-3) node {
$\theta_{i+1}^* = \theta_1$};
\end{tikzpicture}
\caption{}
\end{subfigure} 
\caption{The evolution of $\acst$  for $e_i = 2$ and  $a_i = a_{i+1} = 3$. The hyperparameter $K$ denotes the number of Simulated Annealing steps pre-defined for the generation of elements in  $\acst\low$ and $\acst\hi$. We remark that evolution happens in three different seasons: selection  (a), enlargement (b), and trimming (c and d). \label{fig:dyn_of_ai}}
\end{figure}
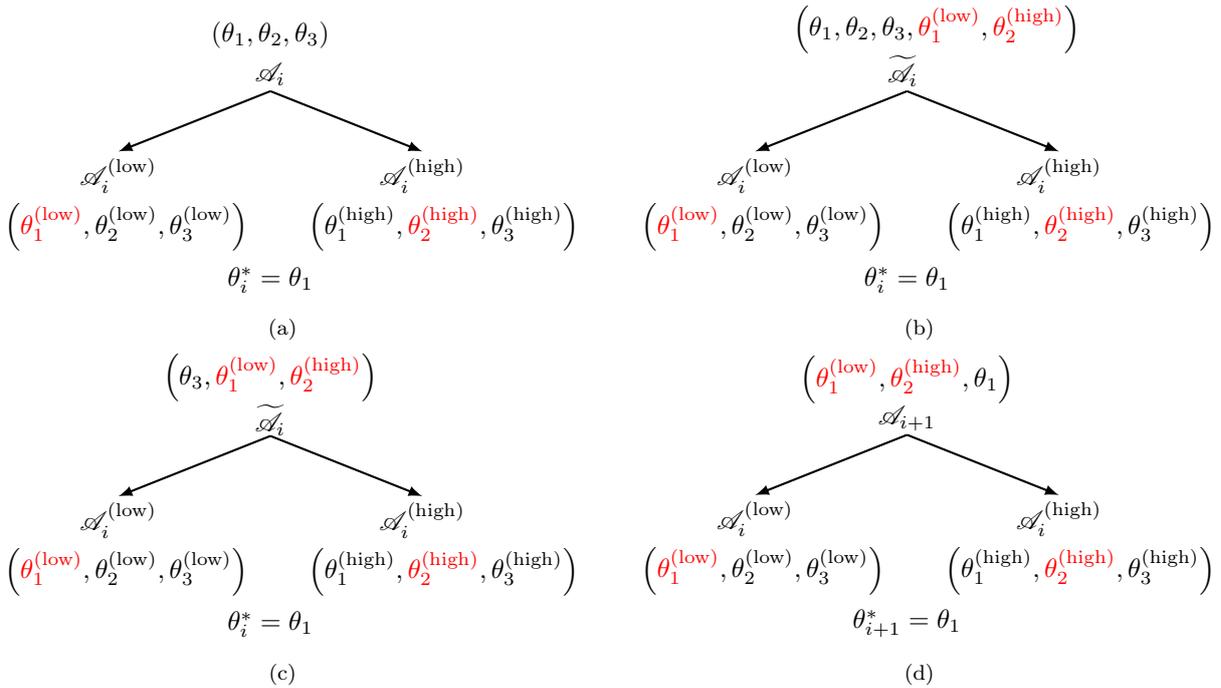

\begin{enumerate}[label= (E\arabic*)]
\item\label{E1} (\textbf{Selection}) Initially, all the elements in $\acst$ evolve by Simulated Annealing with potential  $\mathcal{V}_i(\cdot)$ for $K\low$ steps (resp., $K\hi$), following a policy $T\low$ (resp.,  $T\hi$), yielding a set $\acst\low$ (resp., $\acst\hi$). Then, a predefined hyperparameter $ e_i \in \{1, \ldots,a_i\}$ is used to set the maximum number of evaluations $\theta \mapsto  E(\theta)$ that the model can make, which also sets the number of elements to be chosen from $\acst\low\oplus \acst\hi$. In fact, we split $e_i$ in a random fashion in two parts $e_i\low$ and $e_i\hi$  so that $e_i\low + e_i\hi = e_i$ is satisfied, where  $e_i\low$ states are randomly selected from  $\acst\low$ with policy $\mu\low(\cdot)$, while $e_i\hi$ states are randomly selected from $\acst\hi$ using policy $\mu\hi(\cdot)$; both policies are probability measures over $\acst\low$ and $\acst\hi$, whose construction we clarify in Section \ref{selection_measures}.

\item\label{E2} (\textbf{Enlargement}) The $e_i$ elements chosen above are enqueued to $\acst$ (which, for now, we represent as  $\widetilde{\acst}$) and evaluated by $\theta \mapsto  E(\theta)$ if they are not yet in $\mathscr{H}_i$. In the end, we compute $\theta_{i+1}^* := \mathrm{argmin}_{\theta\in \widetilde{\acst}} E(\theta)$.

\item\label{E3} (\textbf{Trimming}) Elements of $\widetilde{\mathscr{A}}_i$ are dequeued until $a_{i+1}$ elements are left.
If $\theta_{i+1}^*$ is among the dequeued elements  and no copy of it is found in the remaining queue, then it is reinserted into the queue. Then, the process of extraction and  dequeuing is repeated once more to keep $\vert \widetilde{\mathscr{A}}_i \vert = a_{i+1}$; we relabel this queue as $\mathscr{A}_{i+1}$. Furthermore, it holds that $\theta_{i+1}^* = \mathrm{argmin}_{\theta\in \mathscr{A}_{i+1}} E(\theta)$.
\end{enumerate}
Finaly, once $\mathscr{A}_{i+1}$ has been defined, we apply Algorithm \ref{alg:dynamics_Hi} to obtain $\mathscr{H}_{i+1}$. This process  is illustrated in Figure \ref{fig:dyn_of_ai}. 

Several observations are worth making. The first goes back to assumption \ref{H4}, indicating that   $\acst\low$ and $\acst\hi$ are constructed through Simulated Annealing with potential  $\mathcal{V}_i(\cdot)$, allowing for evaluations $\theta \mapsto  E(\theta)$ to be postponed, or rather ``curated'', before actually happening;  in a nutshell and somewhat simplistic way, we can say that ``acceleration'' of convergence happens due to this step, where both the policies $\mu\low(\cdot)$ and $\mu\hi(\cdot)$ play an important role.  One can say that, the trade-off between the amount of imprecision (or ``lack of knowledge'') in the exploration that precedes evaluations $\theta \mapsto  E(\theta)$, and the efficiency in the usage of $\mathscr{H}_i$, is the core of the computational savings that we are looking for.
Further, the invariant on the right of \eqref{property_min} holds, thanks  to \ref{E3}. 

We must also remark that the constructions of both  $\acst$ and $\mathscr{H}_i$ have no parallel in  SA or in  REMC, described  in Section \ref{sec:approaches}, since the evolution in the LSS is driven by branching $\acst$ out as $\acst\low$ and $\acst\hi$ rather than by simple iteration; see Figure \ref{fig:dyn_of_ai}. In that regard, especially if we take into account the policies used in exploration, the sets $\acst\low$ and $\acst\hi$ concern exploration at different spatial scales.
Last, we observe that this multi-agent nature of the LSS adds flexibility to the model, allowing for a large number of  agents to  probe the environment for information gathering at different phases of the exploration.

\subsubsection{A concentration metric.} \label{sub:concentration}
A crucial ingredient in the construction of $\acst$  concerns the quantification  of how concentrated (``clustered'') or well distributed inside $\Omega_i$ its elements are. In fact, this quantity is used to  parameterize several smaller pieces that make up  the LSS, serving as  a proxy for exploration's effectiveness.  Indeed,  if active states are clustered in the same local basin, better exploratory policies should be adopted. This matter gets even more pressing because $\acst$ evolves by branching, which may exacerbate the concentration of active states if left unchecked. On the other hand, whenever active states are not concentrated,  greedier policies that push for minimizing $ E(\cdot)$  should be pursued. 

In the literature of Simulated Annealing, there are several alternative approaches to overcoming the concentration of active states, like restarting the search with different initial active states  or redistributing  active states over the parameter space if they get ``too close to each other''. Yet, this problem can certainly be avoided if, in the long run, a constant evaluation of the spatial concentration of parameters is carried out. 

To avoid the above issues, we devote this section to the  design of such a  measure of concentration,  which we denote as $\mathscr{C}_i \in [0,1]$  and indicates how well distributed  within the box $\Omega_i$  agents are: values of $\mathscr{C}_i$ close to $1$ indicate that particles are becoming clustered; by construction,   $\mathscr{C}_i = 1$ indicates that concentration takes place in a vicinity of  $\theta_i^*$. For now, we explain the mathematics of the measure of concentration $\mathscr{C}_i$ in 1D, generalizing it to higher dimensions later on. For the sake of convenience,  we shall drop any reference to the epoch index $i$: we write  the concentration as $\mathscr{C}$ and the best-known argument at epoch i simply as $\theta^*$. 

At first, consider N agents exploring a box $[a, b]$. We first split the box in $N$ disjoint equal-sized parts $\mathcal{I}_j$, where  $j \in \{1, \ldots N\}$. Afterward,  we count the fraction  of individuals in each of them, stored as $\mu(\{j\})$; clearly,  $\mu(\cdot)$ is a  probability measure.

At first, we compare $\mu(\cdot)$ with two other probabilities, $\nu(\cdot)$ and $\delta_{\theta^*}(\cdot)$; we anticipate that $\nu(\cdot)$ is an uniform measure over bins while $\delta_{\theta^*}(\cdot)$ is concentrated in a single bin. We aggregate these comparisons into a convex combination,
\begin{align}\label{cluster_1D}
\mathscr{C}   = (1 - \lambda) \cdot\mathcal{D}^{(1)}(\mu\Vert \nu) +\lambda \cdot\mathcal{D}^{(2)}(\mu\Vert \delta_{\theta^*}),
\end{align}
of which each term is explained next.  

The quantity $\lambda$ denotes the concentration of  particles in the same bin as the best-known parameter $\theta^*$, and gets adjusted automatically over epochs.

The quantity $\mathcal{D}^{(1)}(\cdot\Vert\cdot)$ is a Kullback-Leiber divergence that compares $\mu(\cdot)$ against $\nu(\cdot)$, an uniform probability measure  over bins, namely,   $\nu(\{i\}) = \frac{1}{N}$. Explicitly, we first define
\begin{align}
\widetilde{\mathcal{D}}^{(1)}(\mu\Vert\nu) = \sum_{j = 1}^N\mu(\{j\})\mathrm{log}\left(\frac{\mu(\{j\})}{\nu(\{j\})}\right).
\end{align}
whose maximum is achieved only when all the agents are concentrated in a single bin. After normalization (so as the maximum value is $1$), we obtain $\mathcal{D}^{(1)}(\mu\Vert\nu)$. 

The second term, $D^{(2)}(\cdot \Vert\cdot)$, works as a sparsity penalization. It compares $\mu(\cdot)$ and the probability measure $\delta_{\theta^*}(\cdot)$, a delta measure concentrated at the bin where $\theta^*$ is (the best explorer to this point).  This is done by using
\begin{align}
\mathcal{D}^{(2)}(\mu\Vert \delta_{\theta^*}) = 1 - \frac{\widetilde{\mathcal{D}}^{(2)}(\mu\Vert \delta_{\theta^*})}{2}, \quad \text{where} \quad
\widetilde{\mathcal{D}}^{(2)}(\mu\Vert \delta_{\theta^*})  = \sum_{j = 1}^N \left\vert \mu(\{j\}) - \delta_{\theta^*}(\{j\})\right\vert.
\end{align}
This finishes the derivation of the metric \eqref{cluster_1D}.  We remark that   $\mathscr{C}$ behaves as a free energy, since its first term is designed  to keep $\mu(\cdot)$ close to $\nu(\cdot)$ - a uniform measure pushing $\mu(\cdot)$ toward  ``thermalization'' - in contrast to its second term, which instead pushes $\mu(\cdot)$ ``far away'' from $\delta_{\theta^*}(\cdot)$ - a measure concentrated on the bin where $\theta^*$  (the first coordinate of $\theta_i^*$) is.\footnote{According to Information theory, both quantities indicate the transportation cost to take one measure to the other. Although the Kullback-Leiber  divergence $\widetilde{D}^{(1)}(\cdot, \cdot)$ is not exactly a metric (due to lack of symmetry), it is useful to quantify the distance between two measures.  The quantity $\widetilde{D}^{(2)}(\cdot, \cdot)$ corresponds to the total variation  norm (or ``Hamming norm''). When the reference measures are the same, both quantities are related, for instance, by the Czizar-Kullback-Leiber inequality  \cite[Chapter 3]{Wainwright}. Nevertheless,  one should note that the first measure scales proportionally to the domains` embedded dimension, while the latter measure is always bounded; cf. \cite[Chapter 9.3.3]{Villani}. This fact, along with the possible variation in the number of active states over epochs, is  the main reason why we normalize $D_1(\cdot, \cdot)$.}

We finally generalize this metric to high-dimensional domains through tensorization across each dimension. Indeed, recall that at the epoch $i$, the best known argument is refereed as $\theta_i^*$ and the domain is a box $\Omega_i = \times_{j = 1}^N [a_j, b_j]$. Now,  we define a concentration metric $\mathscr{C}_{i,j}$ as in \eqref{cluster_1D}  by constructing reference measures $\nu_j(\cdot)$ along the interval $[a_j, b_j]$ and $\delta_{(\theta_i^*)_j}(\cdot)$, where, $(\theta_i^*)_j$ denotes the $j$-th coordinate of $\theta_i^*$. Then, the multi-dimensional concentration metric is defined either as the maximum over the concentration at each dimension, $\displaystyle{\mathscr{C}_i = \max_{1 \leq j \leq N}\mathscr{C}_{i,j}}$, or as the average of these quantities, $\displaystyle{\mathscr{C}_i := \sum_{j=1}^N\frac{\mathscr{C}_{i,j}}{N}}.$

It is known that high dimensional uniform sampling is biased toward high concentrations  due to the ``curse of dimensionality''. In that case, it is expected that most active states cluster on the boundaries. For that reason, one needs to add a parameter that deflates such evaluations dynamically, which the LSS achieves by early stopping: a patience parameter $p_{impr}$ accounts for a minimum number of epochs required for the models to show any improvement, otherwise the search stops. If the number of iterations without improvement exceeds $p_{impr}$, then $\mathscr{C}_i$ gets deflated by a predefined factor, otherwise it gets inflated upwards by a predefined factor toward the value computed in \eqref{cluster_1D}.\footnote{This strategy is acceptable in the field of ML, but lacks Physical motivation. In addition, it may fail in high-dimensional state spaces. A possibly better approach, which somehow generalizes our 
construction, is that of using a different reference measure, constructed as follows. Initially, consider a partition $\left\{I_k\right\}_{k \in \mathcal{B}}$ on  $\Omega_i$, that is, $I_j \cap I_l = \emptyset$, $\displaystyle{\cup_{k \in  \mathcal{B}}I_k = \Omega_i}$. Now we construct a variation of the Gibbs  distribution \eqref{Gibbs},
\begin{align}\label{Gibbs_var}
 \nu_{\text{var}}(I_k) = \frac{e^{\frac{1}{T} \text{dist}(I_k, \partial\Omega_i)}}{Z},
\end{align}
that accounts for the distance between elements in the partition and the boundary $\partial \Omega_i$ of $\Omega_i$. In the limit $T\to \infty$ this measure converges (in an appropriate sense) to the uniform measure $\nu(\cdot)$ used by $\widetilde{\mathcal{D}}_{i,1}^{(1)}(\cdot\Vert\cdot)$. We observe that in  \eqref{Gibbs_var} the temperature $T$ is a hyperparameter  that would have to be adjusted over epochs, a requirement that is dropped by working in the limiting case $T \to \infty$, namely, with the measure $\nu(\cdot)$.}

\begin{Remark}[Computational and Physical aspects of concentration] The spatial concentration of active agents can also be investigated by the introduction of   an interaction potential; for instance, a potential of the  Lennard-Jones type, displaying attraction and repulsion between particles (which, in our case, are agents). This yields an energy space formalism, where spatial concentration is summarized as a 1D measure; in contrast, $\mathscr{C}_i$ quantifies concentration in the state space. 

Computationally, assuming that  $\vert \acst\vert = N$,  the LSS approach requires  $\mathscr{O}(N)$ computations to calculate the concentration of agents (because it only computes the position of active states relative to the best known minimum $\theta_i^*$), whereas the approach using potentials of interaction requires  $\mathscr{O}(N^2)$ computations (because it computes every pairwise distance between elements in $\acst$). These differences are more significant for large values of $N$; it would be interesting to consider and compare different approaches to quantify concentration.
\end{Remark}

\subsubsection{Varying the temperature dynamically -- lowering energy barriers}\label{sec:temperature}

As emphasized before,  high energy barriers may prevent agents from moving by trapping them around metastable states. Dynamic  temperature adjustment is useful to prevent such issues, a technique sometimes referred to as  \textit{lowering energy barriers}. To apply this idea, estimating the amplitude of the barriers seems important. This approach is well known in the literature and has been used in similar contexts \cite{morishita_free}; in some situations, this technique applies more effectively when combined with nonlinear scalings of the energy landscape \cite{morishita_extended}.

In the LSS approach, we estimate the size of the energy barriers as
\begin{align}\label{amplitude}
\alpha_i = \underset{\theta, \theta' \in \mathscr{H}_i}{\max}\,\left( E(\theta)-  E(\theta') \right) = \underset{\theta \in \mathscr{H}_i}{\max}\, E(\theta) -  E(\theta_i^*).
\end{align}
This estimator has the advantage of being  history-dependent and easily computed, with the caveat that  it may possibly overestimate the size of energy landscapes if the parameter space $\Omega_i$ has been thoroughly swept. 
We use \eqref{amplitude} to dynamically adjust policy $T_i\hi$ from a cooling schedule $\widetilde{T}_i\hi$ initially defined, adopting instead
\begin{align}\label{new_high_policy}
\frac{1}{T_i\hi} \gets\frac{(1 - \mathscr{C}_i)}{\widetilde{T}_i\hi} + \frac{\mathscr{C}_i}{4\max\{\alpha_i, \ep\}},
\end{align}
where $\ep>$ is a small term to avoid division by zero. The reasoning behind the construction is as follows: when active states are not clustered ($\mathscr{C}_i \approx 0$), the first term in \eqref{new_high_policy} is dominant, and the policy is not affected by $\alpha_i$. On the other hand, when the set active state is concentrated ($\mathscr{C}_i \approx 1$), we make the second term in \eqref{new_high_policy} dominant but proportional to $\alpha_i$. This step is motivated by the second step \eqref{sim_annealing_transition:b} in Simulated Annealing, where we use the variation in the state-value function $\mathcal{V}_i(\cdot)$ adjusted by the inverse of the temperature:  by taking a state $\theta \in \acst$ and its branched state $\theta\hi \in \acst\hi$, one expects to have at most
\begin{align}
\frac{1}{\alpha_i} \left(\mathcal{V}_i(\theta_i) - \mathcal{V}_i(\theta_i\hi) \right) \approx 1.
\end{align}
Thus, adopting \eqref{new_high_policy} yields a higher  probability of jumping between metastable states, which is exactly what we need to reduce active states clusterization.

\subsubsection{The selection problem -- multi-armed bandits, gold rush, and sampling over the queue $\acst\low \oplus \acst\hi$}\label{selection_measures}

It is yet to be clarified how  elements from  the queue $\acst\low \oplus \acst\hi$ are selected for evaluation $\theta \mapsto  E(\theta)$. In the sequel, we refer to this as the \textit{selection problem}.

Throughout this session, we write the triplet  $(\theta,\theta\low,\theta\hi)\in \left(\acst, \acst\low,\acst\hi\right)$, meaning that both $\theta\low$ and $\theta\hi$ are obtained through Simulated Annealing with potential $\mathcal{V}_i(\cdot)$ under different policies $T_i\low$ and $T_i\hi$, starting at $\theta \in \acst$. An immediate choice for evaluation $\theta \mapsto  E(\theta)$ is
\begin{align}\label{theta_min}
\widetilde{\theta} = \underset{\zeta \in \acst\low \oplus \acst\hi}{\arg\min}  E(\zeta);
\end{align}
subsequent elements can be chosen similarly, from the lowest to the largest. In principle, we can  generalize and relax this ``greedy'' policy approach by using a  ``multi-armed bandits'' technique with $\ep$-policy, where we select elements  in $\acst\low \oplus \acst\hi$ in a probabilistic fashion as to assign a higher probability to lowest values of $ E(\theta)$. However, we readily rule out this possibility because (i) mobility of active states implies that  agents can  hop between local optimum basins and get clustered; in addition to that, (ii)  branching   exacerbates accumulation, resulting in redundancy or crowded exploitation, that is, if we always choose the same element $\widetilde{\theta}$ in \eqref{theta_min}, in subsequent epochs the set of active states $\acst$ will be filled with elements in a neighborhood of $\widetilde{\theta}$. In summary: if adopted indiscriminately, any multi-armed strategy is doomed to failure. Therefore, a different idea, where each agent is stimulated to explore different regions of $\Omega_i$, should be pursued.

In a certain way, the strategy we embrace is akin to a gold rush, where the first to reach a mining site takes all, while others are left with nothing yet are free to explore  and find gold  elsewhere. By nature, this approach parallels the usual exploitation-exploration trade-off in RL, but is better summarized as a \textit{competition-cooperation} trade-off. An analogous phenomenon can be seen in fish schooling due to limited food resources: individuals cooperate as a group while looking for food, but end up competing once finding it; cf. \cite[Section 2]{PSO_kennedy}. Likewise, in the LSS active agents do not compete against each other, but rather share information. In fact, we are interested in agents' accumulated knowledge of the environment, not their individual values. 

In order to evolve and accelerate minimization without resorting to evaluations $\theta \mapsto  E(\theta)$, we must develop a policy that alternates between selections from  $\acst\low$ or $\acst\hi$, sampling is carried out in two steps: as discussed in \ref{E2}, given that  $e_i$ elements must be evaluated $\theta \mapsto  E(\theta)$, $e_i\low \in \mathbb{N}$  of them will follow policies $T\low$, while $e_i\hi \in \mathbb{N}$ of them will follow  $T\hi$ police, in such a way that $e_i = e_i\low +e_i\hi$. Both quantities are parameterized by the concentration $\mathscr{C}_i$, where
\begin{align}\label{e_low_high}
e_i\hi\sim \text{Binomial} \left(n = e_i, p =  \mathscr{C}_i\right);
\end{align}
due to the constraints, we immediately obtain $e_i\low$. In practice, once parameters $e_i\hi$ and $e_i\low$ are defined, we first sample elements from $\acst\low$, a fact that we will come back to later in this section. Parameterization by $\mathscr{C}_i$ is justified by the fact that small values of $\mathscr{C}_i$ indicate ``well-distribution'' of  active states $\acst$  over $\Omega_i$,  implying that more exploitation should be pursued and policy derived from simulated annealing at low temperatures should be adopted. Hence, more states from  $\acst\low$ should be selected. On the other hand, when $\mathscr{C}_i$ is high (close to $1$), active states   $\acst$ are concentrated, and a policy to reduce clusterization  should be pursued; in that case, as if raising the temperature $T_i\hi$, more elements from the queue $\acst\hi$ should be selected.

Once the number of elements to be drawn from each set is defined,  policies $\mu\low(\cdot)$ and  $\mu\hi(\cdot)$  are used  for selection without replacement from the queues $\acst\low$ and $\acst\hi$, respectively. We detail the construction of each one of these policies next. But before doing so, we need to define some auxiliary nonlinear maps based on a softmax function
$\displaystyle{\mathrm{\text{Softmax}}(x;\eta) := \frac{e^{\eta x_i}}{\sum_j^N e^{\eta x_j}}}$, where  $x= (x_1, \ldots, x_N)\in  R^N.$ They are
\begin{align}\label{nonlinear_maps}
\mathcal{N}\mi(x;\eta_1)= \mathrm{\text{Softmax}}(\overline{x};\eta_1), \quad \mbox{and} \quad \mathcal{N}\ma(x;\eta_2)= \mathrm{\text{Softmax}}(\underline{x};\eta_2),
\end{align}
where  $\displaystyle{\overline{x} = \frac{\vert x -x\ma \vert}{x\ma - x\min + \ep}}$, $\displaystyle{\underline{x} = \frac{\vert x -x\mi \vert}{x\ma - x\mi + \ep}}$,  $\displaystyle{x\mi = \min_{1\leq i \leq N}(x_i)}$,  and $\displaystyle{x\ma = \max_{1\leq i \leq N}(x_i)}$;  a small factor $\ep$ avoids division by zero.

Both parameters $\eta_1$ and $\eta_2$ control the distribution of mass over different states, yielding  probability measures from any given vector $x\in \R^N$. In this paper we shall always take $\eta_1<0$ and $\eta_2 <0$, which imply that the following properties hold:
\begin{align*}
\underset{1\leq i \leq N}{\arg\max}\, \mathcal{N}\mi(x;\eta_1) = \underset{1\leq j \leq N}{\arg\min}\, x_j, \quad \mbox{and}\quad
\underset{1\leq i \leq N}{\arg\max}\, \mathcal{N}\ma(x;\eta_2) = \underset{1\leq j \leq N}{\arg\max}\, x_j.
\end{align*}
We are ready to explain both policies. First recall that $\theta\low$ ``branches out'' from $\theta$ through Simulated Annealing with potential $\mathscr{V}_i(\cdot)$. Then,  the policy $\mu\low(\cdot)$ is given as
\begin{align*}
\mu\low (\theta\low) =  \mathcal{N}\mi\left(\mathscr{C}_i \left(\mathcal{V}_i(\theta\low) -  E(\theta)\right) + (1 - \mathscr{C}_i) \mathcal{V}_i(\theta\low);\eta_1\right).
\end{align*}
Intuitively, whenever active states $\acst$ are clustered ($\mathscr{C}_i \approx 1$), the policy $\mu\low (\cdot)$ is close to $$\displaystyle{\mu\low (\theta\low) \approx  \mathcal{N}\mi\left(\mathcal{V}_i(\theta\low) -  E(\theta); \eta_1\right)},$$   favoring elements that show substantial variation at state-value at $\mathcal{V}_i(\theta\low)$ when compared to $ E(\theta)$ evaluated at $\theta$, the state that it branched out from. On the other extreme, $\mathscr{C}_i \approx 0$ indicates that the set of active states $\acst$ is not clustered, therefore a greedier selection seems more appropriate. In this case,  we use $\mathcal{V}_i(\cdot)$ as a ranking proxy, favoring lower values of  $\mathcal{V}_i(\theta\low)$. This establishes the construction of $\mu\low(\cdot)$.

Now we focus on the policy  $\mu\hi(\cdot)$, used to sample elements from $\acst\hi$. Recall that this policy concerns exploration at high temperatures, so a lot of exploration in the active states should be fostered. Initially, one defines a measure of  horizontal displacement $\mathbb{H}_i(b) = \vert b - \theta_i^*\vert^2$ between elements $b \in  \acst\hi$ and $\theta_i^*$ (see Equation \ref{wishful_relaxed}). Then, two quantities are defined: the first  is defined as
$$\ P_{\text{horizontal}}\hi(\theta\hi)\propto \mathbb{H}(\theta\hi),$$
up to a normalization factor that yields a probability distribution that favors values away from $\theta_i^*$.\footnote{The idea has been used in other contexts, for instance,  in the choice of initial points in k-means such that they are as distant from one another as possible; cf. \cite[Section 2.2]{k_means}.}  The second of them is also a probability measure, $\ P_{\text{vertical}}\hi(\cdot)$, parameterized by agents' vertical displacement,
\begin{align}
\ P_{\text{vertical}}\hi(\theta\hi) = \mathcal{N}\ma\left( \vert \mathcal{V}_i(\theta\hi) - \mathcal{V}_i(\theta)\vert; \eta_2\right).
\end{align}
This probability measure is used for sampling from the queue $\acst\hi$ in a biased fashion, favoring agents that display high exploration in amplitude, regardless of the direction it goes. 
At the end of the day, both probability measures are combined as
\begin{align}\label{mu_high_prep}
\widetilde{\mu\hi}(\theta\hi) = (1 - \mathscr{C}_i)\cdot\ P_{\text{vertical}}\hi(\theta\hi) + \mathscr{C}_i\cdot\ P_{\text{horizontal}}\hi(\theta\hi), 
\end{align}
Before defining $\mu\hi(\cdot)$ from  $\widetilde{\mu\hi}(\cdot)$ a preprocessing step is required. Recall  from \eqref{e_low_high} that elements from $\acst\low$ are sampled first. Since elements in the triplet $\left(\acst, \acst\low,\acst\hi\right)$ are linked through branching, for each element $\theta\low \in \acst\low$ that we sample, we remove its corresponding pair in $\acst\hi$; let's call the remaining set $\widetilde{\acst\hi}$ for now. Once this step is performed, $\mu\hi(\cdot)$ is defined in one of the two ways: if $\widetilde{\mu\hi}\Big\vert_{\widetilde{\acst\hi}}(\cdot)$ is  a non-null measure, it is renormalized (so as to be a probability measure) and assigned to $\mu\hi_{\widetilde{\acst\hi}}(\cdot)$ (and zero in the complement), otherwise $\mu\hi(\cdot)$ is simply a uniform measure over $\widetilde{\acst\hi}$ (and zero in the complement). This establishes the construction of $\mu\hi(\cdot)$.

\section{Applications}\label{sec:examples}

The applications we have in sight concern the minimization of functions plagued with local minima, serving as   prototypes of  rugged energy landscape commonly seen in Biophysical Systems and Material Sciences \cite{wales2003energy}.

\begin{figure}[htbp]
\centering
\includegraphics[trim=0cm 0cm 0 0cm, width=.9\textwidth]{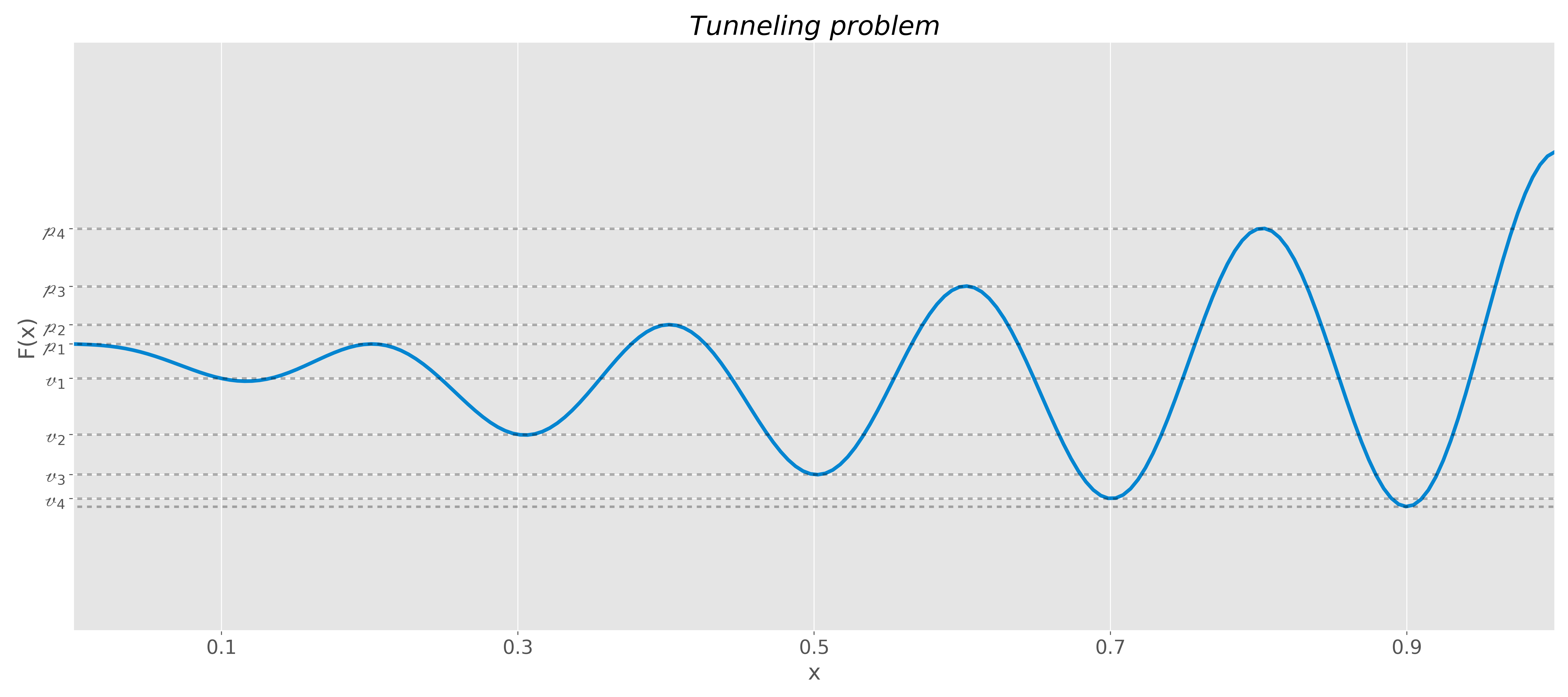}
\caption{The tunneling toy problem in 1D, which shows an oscillatory behavior that is bounded by two curves, $\displaystyle{u(x) = \frac{25 + 30(x-0.1)^2}{25}}$ (top) and $\displaystyle{l(x) =\frac{5 + 25(x-0.9)^2}{25}}$ (bottom). Precisely, $\displaystyle{F(x) = \frac{(1+\lambda(x))}{2}\cdot u(x) +  \frac{(1-\lambda(x))}{2}\cdot l(x)}$, where $\lambda(x) = \sin\left(10\pi x + \frac{\pi}{2}\right)$. By construction, the global  minimum of this function takes place at $x = 0.9$. \label{fig2:tunneling}}
\end{figure}

We shall explore a particular example of a function $G^{(N)}(\cdot): \R^N \rightarrow \R$,
\begin{align}\label{high_dimensional}
G^{(N)}(x_1,\ldots, x_N) = \Pi_{j=1}^N F(x_j),
\end{align}
where $F(\cdot): \R \to \R$ is a real-valued function as depicted in Figure \ref{fig2:tunneling} and whose main properties are discussed next; we shall study this problem in dimension $N\in \{1, 2, 4, 8\}.$

By design, the function $F(\cdot)$  has an abundance of local minima (``valleys'')  and local maxima (``peaks''),  
\begin{align}
v_k := F(-0.1 +  0.2\cdot k) \quad \text{and} \quad p_k := F( 0.2\cdot k), \quad \text{for} \quad k \in \{1, \ldots,5 \},
\end{align}
in such a way that
\begin{subequations}
\begin{align}
v_i > v_{i+1}, \quad \text{for} \quad i \in \{1, ..,5\}, \tag{Better rewards}\\
v_i  -  v_{i+1} > v_{i+1}  -  v_{i+2}, \quad \text{for} \quad i \in \{1, ..,3\}, \tag{Diminishing motivation}\\
p_{i} - v_i<p_{i+1}- v_{i+1}, \quad \text{for} \quad i \in \{1, ..,5\}. \tag{Increasingly harder obstacles}
\end{align}
\end{subequations}
As such, this example embodies some of the main difficulties in the field:  the trapping of parameters on local minima and the issue of tunneling; the latter is commonly seen in Physics, where particles do not diffuse due to high energy barriers separating metastable states. Overall, the designs of $F(\cdot)$ and $G^{(N)}(\cdot)$ are such that particles have to overcome tunneling at higher costs, with progressively smaller ``enthusiasm'' as they move from the initial position $x_{\text{init}} = (0.1, \ldots , 0.1)$ towards the global minimizer at $x_{\text{end}} = (0.9, \ldots, 0.9).$
We have picked an initial state at $x_{\text{init}}$ far away from the global minimizer because our goal is to compare different models rather than exploring any space-filling experimental design; cf. \cite{forrester2008engineering}.

Throughout the discussion we benchmark the results of the LSS with several simulated annealing cases with varying size of Brownian step. We have also split the analysis in  two parts, first focusing on the 1D problem, then moving to its high-dimensional versions.
\subsection{The toy problem in 1D}\label{sec:examples:toy_model}
To begin with, we analyze the LSS in comparison with simulated annealing (SA) under different hyperparameters: SA has been considered with different stepsizes - $\frac{L_i}{12.5}$, $\frac{L_i}{25}$, and $\frac{L_i}{50}$ - where $L_i$ is a vector with the length of the box $\Omega_i$ in each direction. The ML models used for nonlinear interpolation of $\mathcal{V}_i(\cdot)$ varied to contemplate Support Vector Regression models (SVR) and  Neural Networks. As explained in Section \ref{comput_effort}, we alternate between different hyperparameters formulations - and even among different ML models - using a multi-armed bandits technique. 

\begin{figure}[htbp]
\centering
\includegraphics[trim=0cm 0cm 0 0cm, width=.8\textwidth]{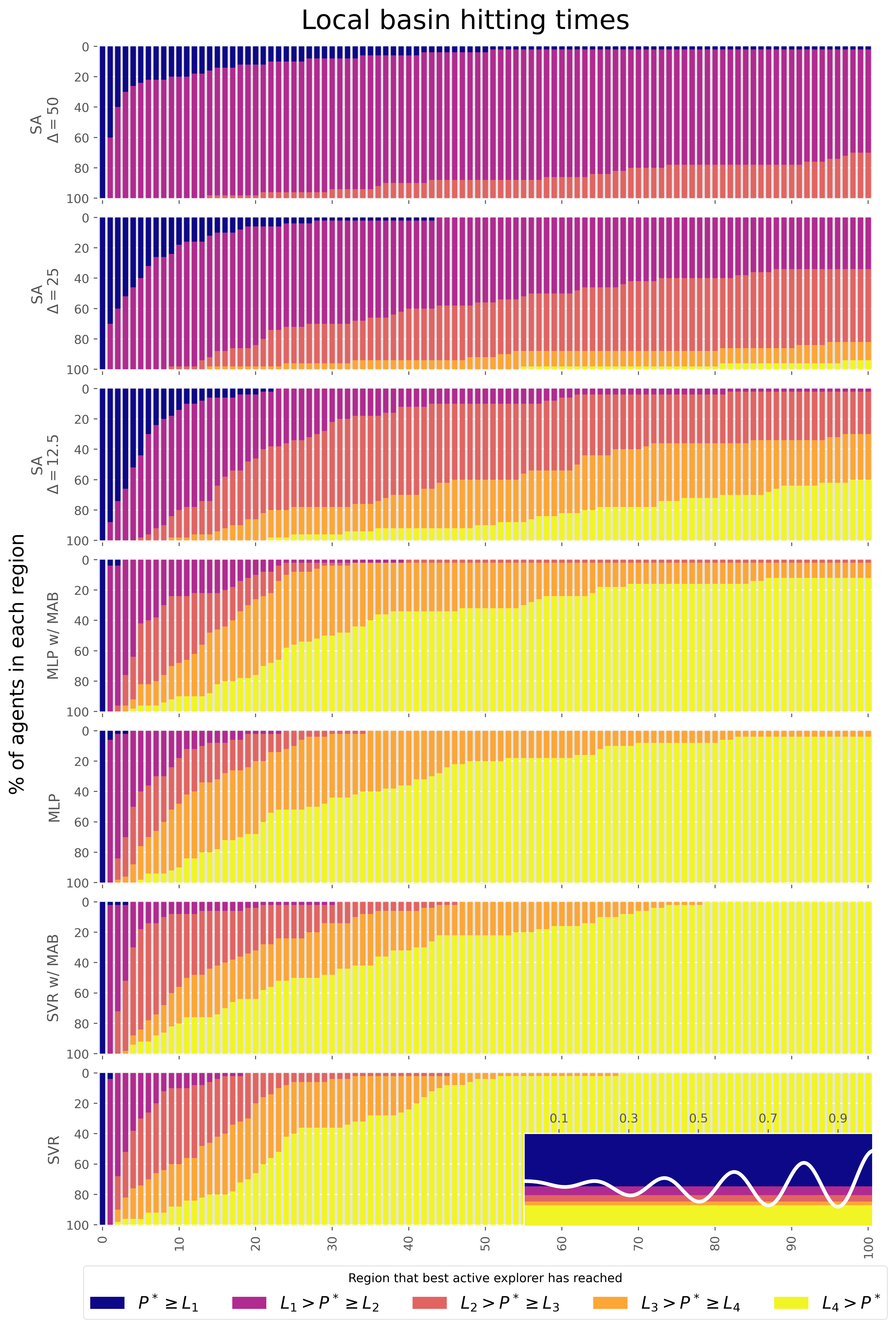}
\caption{Hitting times for different models - SA and LSS. The LSS shows fast minimization, where acceleration varies across different ML models. SA shows a very slow convergence, suffering from the common issues of tunneling and slow diffusion.}\label{All_Hitting_times_sa_versus_lss_June_01_2022.png}
\end{figure}
In Figure \ref{All_Hitting_times_sa_versus_lss_June_01_2022.png} we compare  simulations of different models.  Denoting by $\mathcal{M}_i$ the minimum value of $ E(\theta)$ up to epoch $i$, we study the distribution of  100  simulations  in relation to  level sets of the cost function $G^{(1)}(\cdot)$; observe that this notation is consistent with that introduced in  \eqref{M_functional} for the LSS.
As we can see, simply adjusting the stepsize in Simulated annealing gives a striking improvement in the search. However, the LSS method achieves better results in a much faster way, having a high proportion of agents in the global minimum basin at the end of the experiment. 

Curiously, it is interesting to see that ML models yield very different  acceleration of the optimization, yet their computational cost differ substantially, an opportunity for choosing a ML model that is both efficient and computationally cheap.  Due to this trade-off in model selection, we alternate between two approaches, as explained in Section \ref{comput_effort}.  Among the LSS models studied, one can see that those not using multi-armed bandits seem to perform better. 

\begin{figure}[htbp]
\centering
\includegraphics[trim=0cm 0cm 0cm 0cm, width=.85\textwidth]{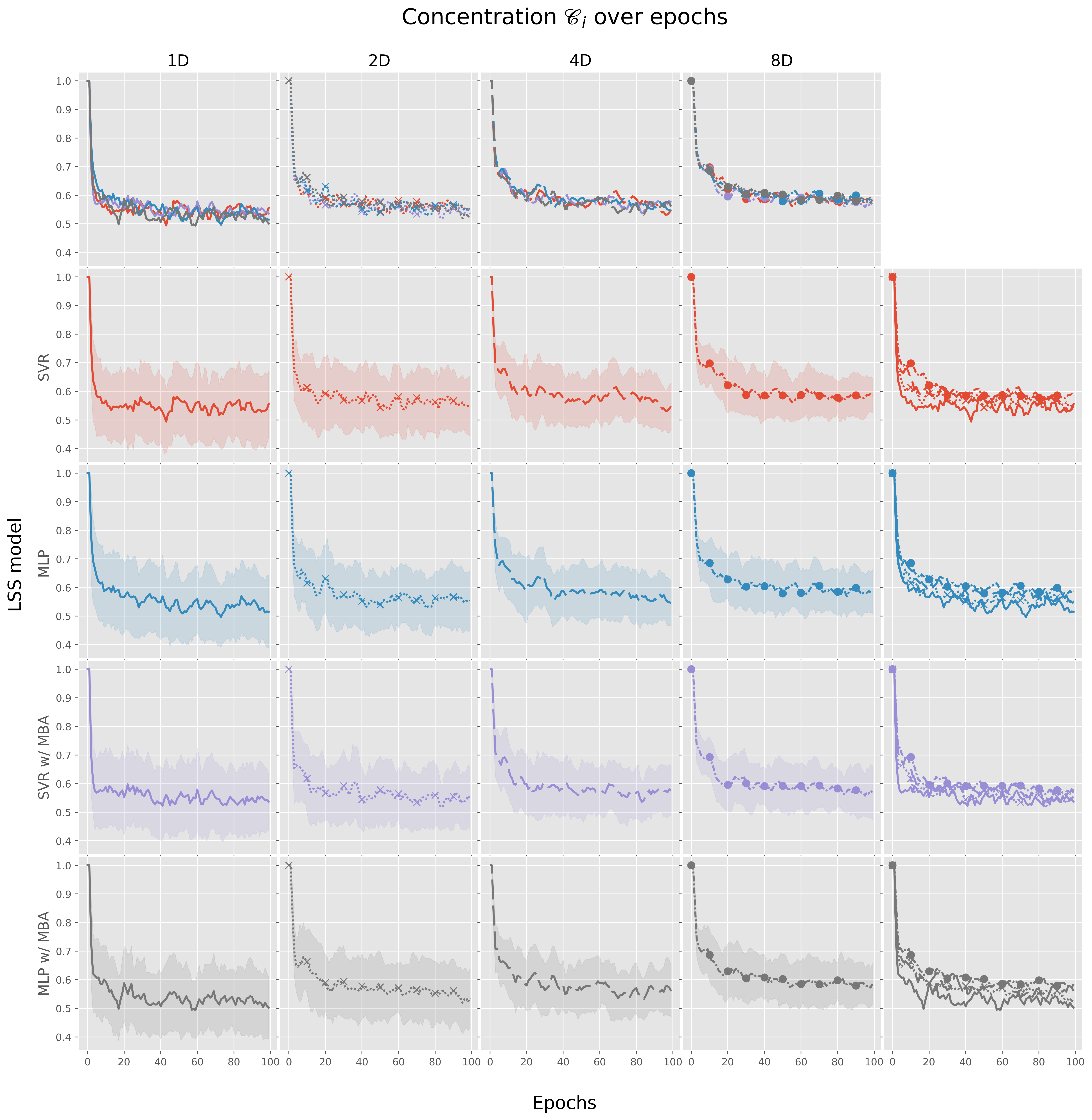}
\caption{Concentration over epochs in different dimensions. It is visible that the high concentration imposed by the initial conditions is quickly ``resolved'' by lowering the concentration. We remark that the figure shows an average behavior, which is smoother than realizations, but it  is expected that realizations are much more oscillatory, since this quantity gets largely affected by the branching of new active agents.  \label{fig:conc_evol}}
\end{figure}

Last, as pointed out in Section \ref{sec:info_sharing}, we keep track of the concentration as the model evolves, as shown in Figure \ref{fig:conc_evol}. Initially, the model starts with three agents concentrated at $x = 0.1$, a state with extremely high concentration (according to the metrics we designed). Particles are then quickly spread out, as shown in the plot, and then concentration remains on average close to 0.5 across the search period; the qualitative behavior of the concentration metric in the  higher dimensions is similar.

\subsection{The toy problem in higher dimensions $N \in \{2,4, 8\}$}
\begin{figure}[htbp]
\centering
\includegraphics[trim=0cm 0cm 0cm 0cm,width=.85\textwidth]{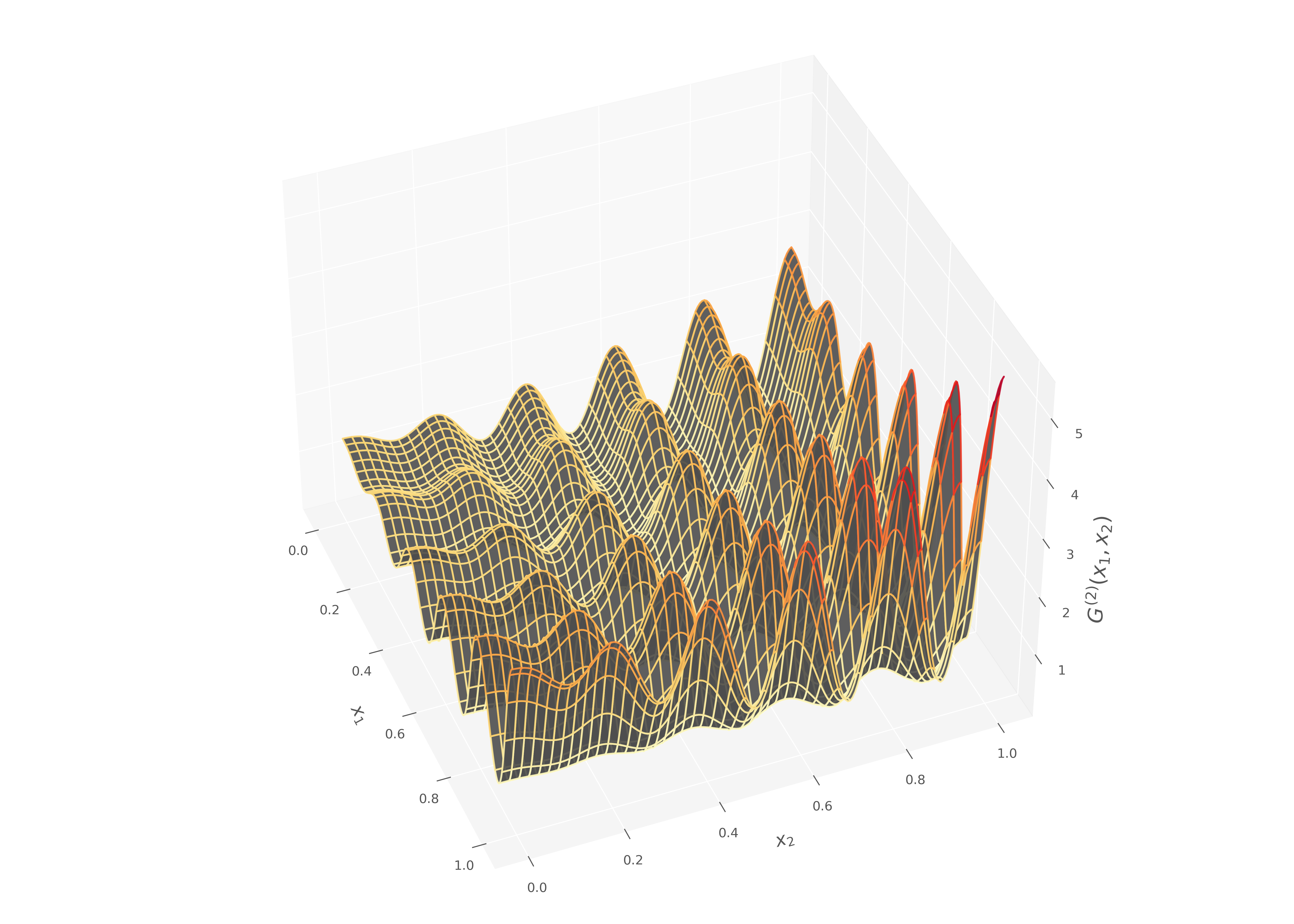}
\caption{The tunneling toy problem in 2D.\label{fig:rugged}}
\end{figure}
In the second part of our applications we focus on studying the role that dimensionality plays in the LSS approach. There is a legitimate concern that agents $\acst$ concentration exacerbates as dimensionality increases, an issue  we aim to curb by the mechanisms described in Section \ref{sec:info_sharing}. Furthermore, by construction,   the cost function $G^{(N)}(\cdot)$  becomes more rugged in high dimensions, and the amount of local minima grows exponentially in $N$, as well as   the number of paths toward the global optimum. 

\begin{figure}[htbp]
\centering
\includegraphics[trim=0cm 0cm 0cm 0cm, width=.8\textwidth]{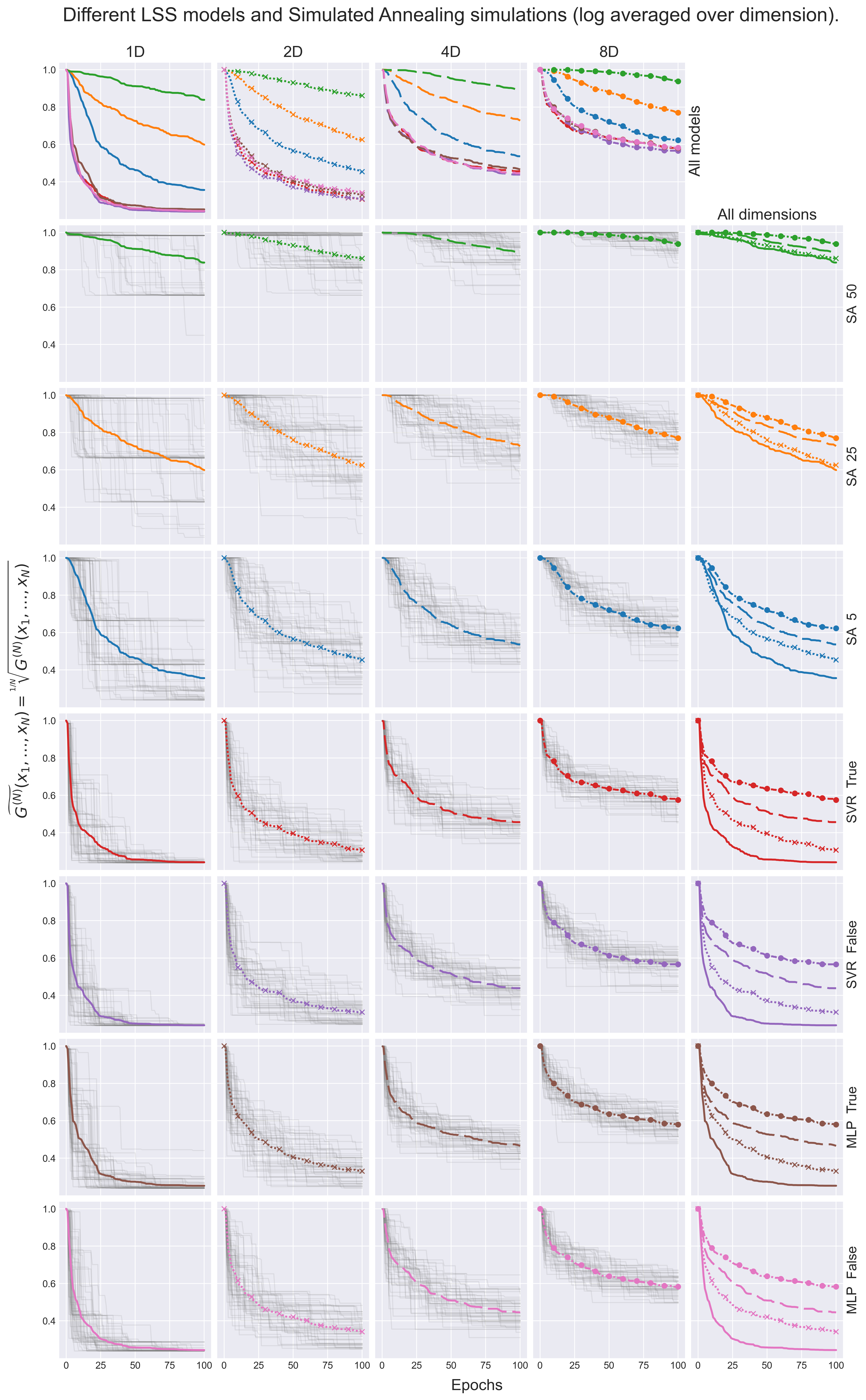}
\caption{The tunneling toy problem in different dimensions for different SA step sizes. Note that the smaller the step size, the more the dimensionality effect gets pronounced, with high-dimensional environments taking a longer time to decay.\label{fig:many_comparison}}
\end{figure}

Plotting decay rates over different dimensions side by side seems reasonable, with a caveat: since  by nature the function $G^{(N)}(\cdot)$ in \eqref{high_dimensional} is a $N$ fold product of $F(\cdot) \in [0,1]$, dimensionality makes values of the cost function closer to zero as $N$ increases. Therefore, we need to correct for dimensionality factors, which we do  by evaluating 
\begin{align}\label{high_dimensional_normalized}
\widetilde{G^{(N)}}(x_1,\ldots, x_N) :	= \sqrt[1/N]{G^{(N)}(x_1,\ldots, x_N)},
\end{align}
instead of $G^{(N)}(\cdot)$. The quantity $\widetilde{G^{(N)}}(\cdot)$ has been  plotted in Figure \ref{fig:many_comparison}, where  one can notice the effect of dimensionality taking place at slowing down convergence towards the minimum. One can also observe that SA's size step  plays an important role in how much progress the model makes. Yet, the LSS shows faster progress, for all the hyperparameter variants considered.

None of the LSS models seem to be faster for all tested dimensions or over epochs. That raises the possibility of still another application of model selection techniques over time; although our code does not contemplate this, such an implementation should be straightforward.

Finally, it can be said that the techniques used to curb concentration are too rigid in low dimensions, making the model spend too much time exploring at high temperatures when it could instead spend more epochs and evaluations at low temperatures. While this approach is welcome in high dimensions, it seems to have a backlash in low-dimensional cases. Nevertheless, in all the cases studied, the LSS approach  accelerates minimization.
\section{Discussion and conclusion}\label{sec:discussion}

We have developed a new global optimization heuristics, the Landscape Sketch and Step approach (LSS), which has been applied in  a toy problem plagued with local minima to mimic rugged energy landscapes. The investigation has been restricted box-shaped domains embedded in low-dimensional spaces with dimensions 1, 2, 4, and 8, focusing on cases where cost function evaluation $\theta \mapsto  E(\theta)$ is high, an aspect that has been key to the architecture of the LSS. To overcome the latter costs, we judiciously substitute information about $ E(\cdot)$ by values of an interpolated function $\mathcal{V}(\cdot)$.

As pointed out, among its many roles, $\mathcal{V}(\cdot)$  works as a state-value function that points out regions in $\Omega$ to be probed for posterior measurements $\theta \mapsto  E(\theta)$. In that regard, it is worth highlighting the parallels and contrasting differences between our approach and that of the method of Particle Swarm Optimization (PSO) \cite{PSO}: PSO is a metaheuristics used for global optimization in which several parameters evolve simultaneously, interacting among themselves in order to minimize a cost function. It is a direct search over the parameter space inspired by phenomena seen in Nature, as fish schooling and bird flocking \cite{PSO_kennedy}. Like all the methods described in Section \ref{sec:approaches}, it also overlooks the cost of evaluation $\theta\mapsto  E(\theta)$ and, by performing searches directly on the parameters space,  does not need to make inferences. In these last two aspects, it  differs greatly from the LSS approach. 

Several remarks are needed in regards to Surrogate Optimization methods. 
Although surrogate modeling using ML models is not new \cite[Chapt. 2]{forrester2008engineering}, the LSS differs from classical approaches in a crucial way:  we do not construct a surrogate model for $ E(\cdot)$, focusing instead on designing $\mathcal{V}_i(\cdot)$, that plays the role of a merit function. Still, the latter is not used in a deterministic fashion, but rather to design a sampling mechanism for candidate points  $\acst\hi \cup \acst\low$ generated by SA; the authors are not aware if this approach is new. The main reason we pursue this approach is the multi-agent nature of the exploratory dynamics, carried out by active states $\acst$. Interestingly, we see similarities in our  construction in \eqref{mu_high_prep} to that in \cite[Equation 1]{wang2014general}, where  a merit function is designed by the convex combination of a surrogate function and the distance to previously evaluated points (as to enforce exploration). Yet, there is a caveat: while their convex combination is performed by cycling weights over a predefined list of values in the interval $[0,1]$, the LSS allows for weights to adjust dynamically by taking the spatial distribution of active states over $\Omega$ into account.

It is clear that a  history-based approach to construct $\mathcal{V}_i(\cdot)$ in an automated fashion has benefits, but it also drawbacks: using $\mathcal{V}_i(\cdot)$ to extrapolate or generalize to  regions not yet seen can be  problematic, especially when $\mathscr{H}_i$ is ``small''; moreover, the LSS approach does not aim at reconstruction of energy  potential functions, which can be attained by  methods like Metadynamics \cite{Laio_2008} or Surrogate Optimization \cite{forrester2008engineering}. Nevertheless,  it must be noted that at the end of the day one can use the sampled points history $\mathscr{H}_i$   as a mean to  reconstruct $ E(\cdot)$ through interpolation, minding that the quality of reconstruction can be difficult to measure unless more information about $ E(\cdot)$  is provided.

A few words about the motivation behind \ref{H4} may illuminate this assumption, which is reasonable in many contexts, especially when $\Delta$ is a finite set that is ``not too large''.  It has been introduced to accommodate cases like \textit{ab initio} calculations and Molecular Dynamics, where computations may take hours to be finalized, usually at a high computational cost (like MD simulations performed on multi-core supercomputers). The problems we had in mind concern the optimization of parameters in \textit{ab initio} calculation, where calculations are much more expensive than classical computations; comparatively, the former plays the role of $\theta \mapsto  E(\theta)$, while the latter plays the role of $\theta \mapsto \mathcal{V}(\theta)$. Iteratively until a halt,  parameters could be optimized at a classical model and, every $k$-steps, evaluation of their fitness could be assessed by the \textit{ab initio} model.

As seen in Section \ref{sec:temperature}, by analyzing the dynamics of agents in the queue of active states one can adjust policy parameters. In the RL literature, this type of interaction is commonly seen in actor-critic methods; It would be interesting to investigate if the application of more advanced policy gradient methods would be more effective in overcoming issues with large energy barriers; cf. \cite{schulman2017proximal}.

A comparison between the LSS approach and the  DYNA method in Reinforcement Learning should be made. The DYNA model  \cite[Chapter 8]{RL}  was developed to adjust action-value functions dynamically in order to accommodate  temporal changes in the environment that an agent explores. On the other hand, the LSS approach is applied in a context where the agent interacts with an  environment that remains the same, yet the agent's  partial knowledge about the environment grows  over time. Nevertheless, the role played by $\acst\low$ and $\acst\hi$ in the LSS approach are somehow comparable to that of sampling from a priority queue in the DYNA method and may be equivalent asymptotically as the number of epochs grows; it would be interesting to further investigate this assertion or implement the LSS approach using ideas of the DYNA method.

To the authors, it seems that the LSS is just an example of a larger class of models that can adopt a similar strategy, namely, that of using an auxiliary function $\mathcal{V}(\cdot)$ modeled on gathered data prior to expensive evaluations $\theta \mapsto  E(\theta)$. For instance, one could apply the same technique to REMC.  The authors also believe that the usage of a detailed balance condition to mix policies and enhance exploration seems more  promising than using probabilistic sampling over queues for the selection of new active states.

Last, we must say that we have not explored any type of halting conditions besides early stopping or putting simply putting a cap on the number of evaluations $\theta \mapsto  E(\theta)$, fixed in advance as a ``budget''.

\subsection{Open problems}
Several directions of investigation could not be contemplated in this paper and are left to future work. 
The biggest among them is proving almost sure convergence of the algorithm. Even though this seems straightforward if the queue of active states $\acst$ has no bounds on its size (that is, $\acst =\mathcal{H}_i$, and classical SA arguments apply), the fact that $\acst$ chops off some of its elements may complicate the proof and, we suspect, in cases of very small length $\acst$ may prevent almost sure convergence from happening.

A few points of improvement seem clear: concentration should either be a vector (amounting for each dimension), or a tensor, containing even more information (average curvature along directions, etc.). 
One could also expand the number of policies beyond $T\low$ and $T\hi$, selecting a few among them over time in order to accelerate convergence. 
With regards to the exploratory behavior and the concentration of agents in $\Omega$, one could explore other methods to construct concentrations $\mathscr{C}$ as in Section \ref{sub:concentration} by using ideas from optimal transportation, as those of used in Stein Variational Gradient Descent, where  the transport of particles is aimed  at finding regions of high probability whereas, at the same time, a repulsive term prevents them from clustering altogether; see \cite[Equation 8]{liu2016stein}. Their construction seems to be more general than ours, and could yield a way to drive active states without the need for computing active states concentration.

If further information about the cost function $ E(\cdot)$ is provided, like its sensitivity to parameters (in the form of higher derivatives), a more careful box shrinking schedule could be devised. In fact, by sampling a few points one could prune subregions in $\Omega$ where the minimum cannot be. We consider such  scenarios unrealistic in real applications, unless the cost function's complexity is very low, like that of linear functions. In such cases, more effective and well-studied techniques are available, like  Branch and Bound methods. Still, similar techniques to the  latter have been used in some proprietary software for better exploration of subproblems derived from branching; cf. \cite{MathWorks}.	

\subsection{Notes and remarks}

\subsection*{Credit authorship contribution statement}

Rafael Monteiro: Conception, Initial discussions, Analysis, Manuscript preparation, Computational Modeling and coding, Simulations, and Algorithm construction. 

Kartik Sau:
Conception, Initial discussions, Analysis, and Manuscript preparation.

\subsection*{Supplementary Material, Data and Code availability}
All the code for this paper has been written in Python and is available on GitHub \cite{MD_ML_github}. Relevant data is also available on the same repository.

\subsection*{Declaration of competing interest}
The authors declare that they have no known competing financial interests or personal relationships that could have appeared to
influence the work reported in this paper.
\subsection*{Acknowledgments}
R.M. would like to thank T. Nakanishi (AIST / MathAM-OIL) for several discussions about Statistical Physics.  He is also grateful to A. Kolchinsky (Universal Biology Institute/The University of Tokyo) for interesting discussions about Information Theory. He is also grateful to T. Morishita (AIST / Tsukuba), who  explained to him some of the underpinnings of Molecular Dynamics simulations and well-established ideas in the literature, besides listening to the description of a preliminary version of the LSS method.

K.S. would like to thank T. Ikeshoji (AIST/ MathAM-OIL) for his encouragement to encounter this problem.

Both authors are very grateful to  N. Yoshinaga (AIST / MathAM-OIL) for carefully reading and commenting on a draft version of this paper, besides a helpful and insightful explanation of the Replica Exchange Monte Carlo methods in Bayesian optimization.

A large community of developers has diligently maintained and improved several libraries - Tensorflow, Numpy, Scikit-learn - that have been used during this project; the authors thank all of them.

\appendix
\section{When is it better to use the  LSS approach?}\label{app:costs}

Counting the number of evaluations using the target  potential is tricky because the model starts up by sampling and using classical simulated annealing until a certain number of samples has been collected.\footnote{This number can be easily adjusted, but we set it as 3, the minimum required number for cross-validations with 3 sets to be performed.} Even though assumption \ref{H4} justifies our investigation, in practice there is an obvious trade-off in the usage of the LSS approach since we have to fit a Machine Learning model at every step, and perform even more interpolations whenever model selection is required.

Although the number of costly evaluations $\theta \mapsto  E(\theta)$ in each epoch $i$ is  at most $e_i$, the number of simulated annealing steps required for acceleration is $K\low + K\hi$; cf. Step \ref{E1}.  This necessity may slow down computations but allows the LSS to perform a better informed search. In the sequel we compare the LSS method with an ``equivalent'' classical Simulated Annealing, in the sense that  all evaluations $\theta \mapsto  E(\theta)$ in the LSS are  not replaced by  fitted potential evaluations $\theta \mapsto \mathcal{V}_i(\theta)$.l Recalling the hyperparameters defined in Section \eqref{sec:info_sharing} -  $K\low$, $K\hi$, $e_i$, $a_i$, the number of evaluations is described below:

\begin{table}[ht]
\centering
\begin{tabular}{|c|c|c|}
\hline
\rowcolor{lightgray} 
\textbf{Type of computation} & \textbf{Classical Simulated Annealing} & \textbf{LSS} \\[1ex]
\hline
$\theta \mapsto  E(\theta)$ & $\displaystyle\sum_{i = 1}^{N^*}(K_i\low + K_i\hi)a_i$ & $\displaystyle\sum_{i = 1}^{N^*} e_i$ \\
\hline
$\theta \mapsto \mathcal{V}(\theta)$ & - & $\displaystyle\sum_{i = 1}^{N^*}(K_i\low + K_i\hi)a_i$ \\
\hline
\end{tabular}
\caption{Number of evaluations required in the LSS when compared to classical Simulated Annealing. Recall from Section \eqref{sec:info_sharing}  that $N^*$ denotes the number of epochs and, at epoch $i$;  $K_i\low$ and $K_i\hi$ represent the number of SA steps required to generate each element in $\acst\low$ and $\acst\hi$, respectively;  $e_i$ indicates the number of evaluations  $\theta \mapsto  E(\theta)$, and finally  $a_i$ denotes the length of the queue $\acst$.}
\end{table}
In a context where fitting costs are negligible, evaluations $\theta \mapsto  E(\theta)$ and $\theta\mapsto\mathcal{V}(\theta)$ have costs $\mathscr{C}_{ E}$ and $\mathscr{C}_{\mathcal{V}}$, respectively, assumption \ref{H4} can be quantified as
$$
\left(\sum_{i = 1}^{N^*}(K_i\low + K_i\hi)a_i\right)\cdot \mathscr{C}_{ E}  \gg \left(\sum_{i = 1}^{N^*} e_i\right)\cdot \mathscr{C}_{ E}  + \left(\sum_{i = 1}^{N^*}(K_i\low + K_i\hi)a_i\right)\cdot\mathscr{C}_{\mathcal{V}}.
$$
\section{Protocols}
Standards to be followed in order to integrate the different programs used in this paper.

\subsection{Toy model experiments}

We compare two types of models:
\begin{itemize}
\item 100 simulated annealing experiments.
\item 100 landscape sketch and step experiments with 3 agents each.
\end{itemize}
The number of box searches and within box searches is 10.

In both cases, the simulated annealing step is the same for deep evaluations. The LSS also has an exploratory phase that has a step size that varies according to the concentration of particles: under a high concentration of agents, the step size is the same as that of deep explorations, otherwise it is half of such a step size when agents are not concentrated.

\section{Methodology and statistical analysis}

\paragraph{Hyperparameter tuning of the fitting models.}

In the Neural Network model there are several network architecture constraints set in advance like the number of hidden layers, regularization rates etc; they are known as \textit{hyperparameters}. Their values can be tuned in order to maximize the performance of the network during its training phase, mostly taking accuracy or loss metric in consideration.  In this work hyperparameter tuning was carried out only for learning rates  and dropout rates, using 3 fold cross-validation, where the train set was split in 3 parts of roughly the same size, followed by 3 evaluations, in each of which one of these sets is held out, while parameters are fitted in the other two. Once the training process is over, the accuracy of the model is measured, and the process is repeated with another piece of these 3 sets held aside. This is carried out until the 3 parts are exhausted, and the evaluation is reached as an average. Further investigation on other model architectures  - number of hidden layers, larger grid search - are also possible but have not been carried out. The code available on GitHub readily allows  such experiments; cf. \cite{MD_ML_github}. 

\section{Computational methodology}

The optimization of the empirical potential parameters involves multi-platform computations concerning different parts of the model. Machine Learning steps were carried out in Python 3.6. 

All the Python code has been built using Tensorflow, a Python library well suited for Machine Learning problems \cite{tensorflow2015-whitepaper}.

%
%


\newcommand{\etalchar}[1]{$^{#1}$}

\end{document}